\documentclass[english,12pt]{article}
\usepackage[T1]{fontenc}
\usepackage[utf8]{inputenc}
\usepackage{mathptmx}
\usepackage[a4paper]{geometry}
\usepackage{babel}
\usepackage{array}
\usepackage{float}
\usepackage{amsmath}
\usepackage{amssymb}
\usepackage{graphicx}
\usepackage{hyperref}
\usepackage{breakurl}
\usepackage{amsthm}
\usepackage{amsfonts}
\usepackage{tikz}
\usepackage{xcolor}
\usepackage{algorithm}
\usepackage{algorithmic}
\usepackage{lscape}
\usepackage{colortbl}
\makeatletter
\makeatletter

\providecommand{\tabularnewline}{\\}

\theoremstyle{definition}
\newtheorem{defn}{\protect\definitionname}
  
\theoremstyle{proposition}
\newtheorem{propo}{\protect\propositionname}
  
\theoremstyle{corollary}
\newtheorem{coro}{\protect\corollaryname}

\theoremstyle{lemma}

\makeatother

\usepackage{babel}
\providecommand{\definitionname}{Definition}
  
\usepackage{babel}
\providecommand{\propositionname}{Proposition}
  
\usepackage{babel}
\providecommand{\corollaryname}{Corollary}  

\usepackage{babel}
\providecommand{\lemmaname}{Lemma} 

\makeatother
\def \ind {{{\rm 1}\hskip-2.2pt{\rm l}}}

\begin{document}
\begin{center}
\textbf{\Large {Classification Approach based on Association Rules
mining for Unbalanced data}}{\Large }\\
{\Large{} \bigskip{}
\bigskip{}
}{\small $\textbf{Cheikh Ndour}^{1,2,3}$ \& $\textbf{Aliou Diop}^{1}$ \& $\textbf{Simplice Dossou Gbété}^{2}$
}\\
{\small{} }
\end{center}{\small \par}
{\small ${}^1$ \textit{Laboratoire d'Etudes et de Recherche en Statistiques et Développement (LERSTAD), Unversité Gaston Berger, Saint-Louis, Sénégal}}
{\small \par}
{\small ${}^2$ \textit{Laboratoire de Mathématiques et de leurs Applications (LMA), UMR CNRS 5142, Université de Pau et des Pays de L'Adour, Pau, France} }
{\small \par}
{\small ${}^3$ \textit{Institut de Santé Publique, d'Epidémiologie et de Développement (ISPED), INSERM U897, Université de Bordeaux, Bordeaux, France} }

\bigskip{}
\begin{abstract}
This paper deals with the binary classification task when the target class has the lower probability of occurrence. In such situation, it is not possible to build a powerful classifier by using standard methods such as logistic regression, classification tree, discriminant analysis, etc. To overcome this short-coming of these methods which yield classifiers with low sensibility, we tackled the classification problem here through an approach based on the association
rules learning. This approach has the advantage of allowing the identification of the patterns that are well correlated with the target class. Association rules learning is a well known method in the area of data-mining. It is used when dealing with large database for unsupervised discovery of local patterns that expresses hidden relationships between input variables. In considering association rules from a supervised learning point of view, a relevant set of weak classifiers is obtained from which one derives a classifier that performs well. 
\end{abstract}

\section{Introduction}
This paper deals with the binary classification task when the target class has the lower probability of occurrence. In such situation, standard methods such as logistic regression \cite{Bscarpa}, classification tree, discriminant analysis, etc. do not make it possible to build an effective classification function  \cite{menardi_training_2014}. They tend to focus on the prevalent class and to ignore the target class. Several works were devoted on this subject, even in the recent past, as well as from the conventional statistical viewpoint as such that of machine learning. Some works among them will consider the improvement of the regression models' fitting to produce a classification function with a small prediction bias without loosing interesting features of the standard methods as the ability to evaluate the contribution of each covariate in the variations of target class probability (regression methods) or the identification of the pattern correlated with the target class(tree method). Alternative approaches consist in aggregation techniques like boosting and bagging \cite{Hualin2007} which combine multiple classification functions with large individual error rate to produce a new classification function with smaller error rate \cite{Breiman96}.

Our aims is to propose a statistical learning method that provides  an effective classifier and allows to identify relevant patterns correlated with the target class. To achieve this goal we took the route toward the association rules learning which is a well known method in the area of data mining. It is used when dealing with large database for unsupervised discovery of local-patterns that express hidden and potential valuable relationships between feature variables. In considering association rules from a supervised statistical learning viewpoint, a relevant set of weak classifiers is obtained from which one derives a classification function that performs well. Such an approach is not actually new since it has been already considered in the machine learning literature \cite{JliKelman2005}. In the present work we aim at inserting it within the traditional framework of the statistics  and showing its relevance by its application to real datasets.

\section{Patterns, Pattern-based binary classifier and association rules}
Let  $X=(X_j)_{j=1:p}$ be a sequence of $p$ random variables where each component $X_j$ is a categorical variable that takes its values $m^{X_j}_{h_j}$ on a nominal scale made of $q_j$ levels. Let denote the domain of $X_j$ by $ Dom(X_j)= \{m^{X_j}_{h_j}; \,h_j=1:q_j\} $. Then the domain of values of $X$ is \,
${\displaystyle Dom(X) = \prod_{i=1}^{n} Dom(X_j)}$. In what follows the notation $\left[ X_j=m^{X_j}_{h_j}\right] $ will denote the event which occurred when $m^{X_j}_{h_j}$ is the value of the variable  $X_j$ as well as the indicator of that event.

\subsection{Pattern}
\begin{defn} A pattern $U$ is an intersection of elementary events $\left[ X_j=m^{X_j}_{h_j}\right] $ where $m^{X_j}_{h_j}$ is a modality of the variable $X_j$ and  $J$ is a subset of $1:p$. It will be denoted by 
\[
U = \bigcap_{j \in J} \left[X_j =m^{X_j}_{h_j} \right]
\]
\end{defn}

In order to simplify the notation we write $\left(m_{h_j}^{X_j}\right)_{j \in J}$ to denote the pattern $U$. From statistical viewpoint a pattern  $\left(m_{h_j}^{X_j}\right)_{j \in J}$ can be understood as the expression of an interaction between categorical variables $\left(X_{j}\right)_{j\in J}$ that it is made of and hence the event ${\displaystyle \bigcap_{j \in J} \left[ X_j = m^{X_j}_{h_j} \right] }$ 
is a relevant pattern if it has a significant probability of occurrence. This probability is called coverage of the pattern $U$. The length of the pattern is equal to the cardinal of the indexes subset $J$. It defines the complexity of the pattern and the greater the number of variables jointly performed the higher the complexity of the pattern is. Therefore the number of observations checking the pattern becomes increasingly small.  With this in mind, we can state some relationship between patterns.

\begin{defn}~\\ 
\begin{enumerate}
\item Two patterns  $\displaystyle \left( m_{h_l}^{X_l}\right) _{l\in L}$ and $\displaystyle \left( m_{h_j}^{X_j}\right) _{j\in J}$ are disjoint patterns if the indexes subsets $L$ and $J$ are disjoint. (i.e\quad $L\cap J=\emptyset$).\\

\item The pattern $\displaystyle \left( m_{h_l}^{X_l}\right) _{l\in L}$  is nested in the pattern  $\displaystyle \left( m_{h_j}^{X_j}\right) _{j\in J}$ if:
\begin{itemize}
\item[a)] $L\subset J$
\item[b)] $\forall \, l \in L, \, \forall h_l \in \{1:q_l\} \quad \exists \,! \, j \in J, \, \exists\,! \,k_j \in \{1:q_j\} \textrm{ such that } \, m_{h_l}^{X_l}=m_{k_j}^{X_j}$ 
\end{itemize}
\end{enumerate} 
\end{defn}
In the field of computer science, an elementary event $\left[ X_j=m^{X_j}_{h_j}\right] $ is called item , while a pattern $U$ is named an itemset. The length of the pattern  
$\displaystyle \left(m_{h_j}^{X_j}\right) _{j\in J}$
 is equal to the size of the set $J\subset 1:p$.
 
\subsection{Example of pattern}
As application data set, consider here  the Adult Data Set that is extracted from the 1994 Census database \cite{kohavi_nbtree}. Adult Data Set mainly contain individuals aged over 16 years old and having both an adjusted gross income greater than 1 and an hourly volume of positive work. Data refer to 45222 individuals without the missing data. 
Data contain 14 covariates which 6 are continuous variables (age, fnlwgt, education number, capital gain, capital loss, hours per week) and 8 are categorical variables (work class, education, marital status, occupation, relationship, race, sex and native country) and one binary response variable (income group) indicating if  the annual income of an individual is over $\$$50K or not. Prediction task is to determine a predictive pattern for which a person makes over $\$$50K a year. A part of this application data set is presented in the Table \ref{tab1} containing the six first records and  the first eleven descriptive variables. 

\begin{table}[ht]
\centering
{\footnotesize
\begin{tabular}{p{2.5cm}p{4cm}p{8cm}}
\hline
Variable & Definition & modality \\
\hline
age       & the individual's age & less than 22, 22-30, 30-38,  38-46,  48-54, 54-62, 62 years and over\\
workclass & the individual's work-class service & Private, Self-emp-not-inc, Self-emp-inc, Federal-gov, Local-gov, State-gov, Without-pay, Never-worked\\
education & the individual 's education level& Bachelors, Some-college, 11th, HS-grad, Prof-school, Assoc-acdm, Assoc-voc, 9th, 7th-8th, 12th, Masters, 1st-4th, 10th, Doctorate, 5th-6th, Preschool\\
marital-status & marital status of a individual & Married-civ-spouse, Divorced, Never-married, Separated, Widowed, Married-spouse-absent, Married-AF-spouse\\
occupation & individual's profession & Tech-support, Craft-repair, Other-service, Sales, Exec-managerial, Prof-specialty, Handlers-cleaners, Machine-op-inspct, Adm-clerical, Farming-fishing, Transport-moving, Priv-house-serv, Protective-serv, Armed-Forces\\
relationship & individual's relationship & Wife, Own-child, Husband, Not-in-family, Other-relative, Unmarried\\
race & individual's race& White, Asian-Pac-Islander, Amer-Indian-Eskimo, Other, Black\\
sex & individual 's sex & Female, Male\\
capital-gain & the amount of an individual's capital-gain & {0}, less than  5000, 5000-10000, 10000 and over \\
capital-loss & the amount of an individual's capital-loss &  {0}, less than 1500, 1500-1750, 1750-1950, 1950-2150, 2150 and over \\
hours-per-week & number of hours worked per week & less than 35 35 - 42,  42 - 50,  50 - 65, 65 and over \\
native-country & individual's native country & United-States, Cambodia, England, Puerto-Rico, Canada, Germany, Outlying-US(Guam-USVI-etc), India, Japan, Greece, South, China, Cuba, Iran, Honduras, Philippines, Italy, Poland, Jamaica, Vietnam, Mexico, Portugal, Ireland, France, Dominican-Republic, Laos, Ecuador, Taiwan, Haiti, Columbia, Hungary, Guatemala,
 Nicaragua, Scotland, Thailand, Yugoslavia, El-Salvador, Trinadad \& Tobago, Peru, Hong, Holand-Netherlands\\
\hline
\end{tabular}
}
\caption{list of all explicative variables of the Adult Data Set without education number and final weight variables}
\label{tab1}
\end{table}

An elementary event is defined as an attribute-value pair. For example, [age = 42-55] is elementary event. A pattern is defined as an intersection of elementary events. For example, the following are two patterns that we can extract from the Adult Data Set.\\~~\\
\begin{tabular}{p{1.65cm}p{3.7cm}}
\textbf{Pattern 1} & of length three \\ 
 &$\bullet$ age \, in \, [48,54]\\
 &$\bullet$ relation = Husband \\
 &$\bullet$  hourpw \, in \, [42,50]\\
\end{tabular} 

 \begin{tabular}{p{1.65cm}p{3.7cm}}
\textbf{Pattern 2} & of length two \\ 
 &$\bullet$ age \, in \, [48,54]\\
 &$\bullet$ relation = Husband \\
\end{tabular}
 
 \begin{tabular}{p{1.65cm}p{4cm}}
\textbf{Pattern 3} & of length three \\ 
 &$\bullet$ education = Doctorate \\
 &$\bullet$ workclass = Private\\
 &$\bullet$  hourpw \, in \, [42,50]\\
\end{tabular} 

\textbf{Pattern 1}  is a pattern with three elementary events.  We say that the pattern \textbf{Pattern 2} is nested in the pattern  \textbf{Pattern 1}  because the last one is the first pattern plus the elementary event  [hourpw = 40-42]. \textbf{Pattern 3} is disjoint to \textbf{Pattern 2} because they are non common  covariates.

\subsection{Association rules}
\begin{defn}

{\small{}Let's consider two disjoint local patterns $U=\left(m_{h_{l}}^{X_{l}}\right)_{l\in L}$
and $U^{\prime}=\left(m_{h_{j}}^{X_{j}}\right)_{h\in j}$. An association
rule is an implication of the form $U\rightarrow U^{\prime}$ meaning
that the probabilities $\Pr\left\{ \left[{\displaystyle \prod_{l\in L}}\left[X_{l}=m_{h_{l}}^{X_{l}}\right]=1\right]\wedge\left[{\displaystyle \prod_{j\in J}}\left[X_{j}=m_{h_{j}}^{X_{j}}\right]=1\right]\right\} $
and }
{\small{}$\Pr\left\{ \left[{\displaystyle \prod_{j\in J}}\left[X_{j}=m_{h_{j}}^{X_{j}}\right]=1\right]\mid\left[{\displaystyle \prod_{l\in L}}\left[X_{l}=m_{h_{l}}^{X_{l}}\right]=1\right]\right\} $
are significant. }{\small \par}
\end{defn}

\begin{enumerate}
\item The  probabilities {\small{}$\Pr\left\{ \left[{\displaystyle \prod_{l\in L}}\left[X_{l}=m_{h_{l}}^{X_{l}}\right]=1\right]\wedge\left[{\displaystyle \prod_{j\in J}}\left[X_{j}=m_{h_{j}}^{X_{j}}\right]=1\right]\right\} $
\linebreak{} and $\Pr\left\{ \left[{\displaystyle \prod_{j\in J}}\left[X_{j}=m_{h_{j}}^{X_{j}}\right]=1\right]\mid\left[{\displaystyle \prod_{l\in L}}\left[X_{l}=m_{h_{l}}^{X_{l}}\right]=1\right]\right\} $
are significance if they exceed specified thresholds.}{\small \par}
\item {\small{}An association rule $U\rightarrow U^{\prime}$ expresses
the fact that not only there is a high probability that the events
}\linebreak{}
{\small{}$\left[{\displaystyle \prod_{l\in L}}\left[X_{l}=m_{h_{l}}^{X_{l}}\right]=1\right]$
and $\left[{\displaystyle \prod_{j\in J}}\left[X_{j}=m_{h_{j}}^{X_{j}}\right]=1\right]$
occur simultaneously but also $\left[{\displaystyle \prod_{j\in J}}\left[X_{j}=m_{h_{l}}^{X_{j}}\right]=1\right]$
has a high probability of occurrence under the conditions specified
by the event $\left[{\displaystyle \prod_{l\in L}}\left[X_{l}=m_{h_{l}}^{X_{l}}\right]=1\right]$. }{\small \par}
\item {\small{}The probability $\Pr\left\{ \left[{\displaystyle \prod_{l\in L}}\left[X_{l}=m_{h_{l}}^{X_{l}}\right]=1\right]\wedge\left[{\displaystyle \prod_{j\in J}}\left[X_{j}=m_{h_{j}}^{X_{j}}\right]=1\right]\right\} $
is called the support of the association rule. It tells us how frequent
the event $\left[{\displaystyle \prod_{l\in L}}\left[X_{l}=m_{h_{l}}^{X_{l}}\right]=1\right]$
and $\left[{\displaystyle \prod_{j\in J}}\left[X_{j}=m_{h_{j}}^{X_{j}}\right]=1\right]$
can occur simultaneously.}{\small \par}
\item {\small{}The conditional probability $\Pr\left\{ \left[{\displaystyle \prod_{j\in J}}\left[X_{j}=m_{h_{j}}^{X_{j}}\right]=1\right]\mid\left[{\displaystyle \prod_{l\in L}}\left[X_{l}=m_{h_{l}}^{X_{l}}\right]=1\right]\right\} $
is called the confidence of the association rule. It is a measure
of the correlation between the events $\left[{\displaystyle \prod_{j\in J}}\left[X_{j}=m_{h_{j}}^{X_{j}}\right]=1\right]$
and $\left[{\displaystyle \prod_{l\in L}}\left[X_{l}=m_{h_{l}}^{X_{l}}\right]=1\right]$.}{\small \par}

\item When dealing with an association rule $U \longrightarrow U'$, the pattern $U$ is named as the right-hand side or the antecedent of the association rule while the pattern $U'$ is called left-hand side or consequence of the association rulrule.
\end{enumerate}

The next subsection will address the relationship between the binary classifier members of special case of association rules.

\subsection{Pattern-based binary classifier and binary association rule}

Let's consider a random pair $\left(X\mbox{,}Y\right)$ where $Y$
is a Bernoulli variable and $X=\left(X_{j}\right)_{j=1:p}$ is a multivariate
random element made of categorical marginals components. 
\begin{defn}
A pattern-based classifier is a function $x\rightarrow\phi\left(X\mbox{,}U\right)\left(x\right)$
defined on the set $\mbox{Dom}\left(X\right)$ of the values of the
random element X such that 
\[
\phi\left(X\mbox{,}U\right)\left(x\right)={\displaystyle \prod_{l\in L}}\left[X_{l}=m_{h_{l}}^{X_{l}}\right]\left(x\right)
\]
with $U=\left(m_{h_{l}}^{X_{l}}\right)_{l\in L}$
\end{defn}

As a classifier a pattern-based binary classifier is characterized by performances metrics such as error rate (err), sensitivity or true-positive rate (tpr), specificity or true-negative rate (tnr), false positive rate (fpr), positive predictive value (ppv) and negative predictive rate.
\begin{enumerate}
\item $\mbox{err}=\Pr\left\{ \phi\left(X\mbox{,}U\right)\neq Y\right\} $
\item $\mbox{tpr}=\Pr\left\{ \left[\phi\left(X\mbox{,}U\right)=1\right]\mid\left[Y=1\right]\right\} $
\item $\mbox{tnr}=\Pr\left\{ \left[\phi\left(X\mbox{,}U\right)=0\right]\mid\left[Y=0\right]\right\} $
\item $\mbox{fpr}=1-\mbox{tnr}$
\item $\mbox{ppv}=\Pr\left\{ \left[Y=1\right]\mid\left[\phi\left(X\mbox{,}U\right)=1\right]\right\} $
\item $\mbox{npv}=\Pr\left\{ \left[Y=0\right]\mid\left[\phi\left(X\mbox{,}U\right)=0\right]\right\} $
\end{enumerate}

If the class distribution of the response variable is unbalance and the probability of the target class is very small one may encounter very often pattern-based binary classifier $\phi\left(X\mbox{,}U\right)$ that have low true-positive rate. For example, the  following are two patterns that we observe after unbalancing the adult Data Set where we set the prevalence class proportion at $\alpha = 0.7\%$. \\

\begin{tabular}{p{1.65cm}p{5cm}cccccc}
\textbf{Pattern 4} & of length three & \multicolumn{6}{c}{ \textbf{pattern-based binary classifiers's performances}}\\
 &$\bullet$ workclass= Local-gov & tpr & tnr & fpr & err & ppv & npv \\ 
 &$\bullet$ minority.group= White &0.062 &0.962 &0.031 &0.037 & 0.014 & 0.993\\
 &$\bullet$  hourpw \, in \, [35,42] & & & & & \\
 
\end{tabular}

\begin{tabular}{p{1.65cm}p{5cm}cccccc}
\textbf{Pattern 5} & of length three & \multicolumn{6}{c}{ \textbf{pattern-based binary classifiers's performances}}\\
 &$\bullet$ relation= Husband & tpr & tnr & fpr & err & ppv & npv \\
 &$\bullet$ cgain= 0 &0.185 &0.939 &0.054 &0.059 & 0.023 & 0.994\\
 &$\bullet$  hourpw \, in \, [50,65] & & & & & \\
\end{tabular}

The pattern-based binary classifiers associated to \textbf{Pattern 4} and \textbf{Pattern 5} are weak classifiers.\\

Since the true-positive rate increases  as the value of $ \Pr\left\{ \left[\phi\left(X\mbox{,}U\right)=1\right]\wedge\left[Y=1\right]\right\}$ increases, one should higher attention to patterns for which $ \Pr\left\{ \left[\phi\left(X\mbox{,}U\right)=1\right]\wedge\left[Y=1\right]\right\}$ exceeds some specified threshold $s_0 \in \left]0, \Pr\left\lbrace \left[Y=1 \right]  \right\rbrace  \right] $. Moreover  the error rate is equal to : \linebreak $\Pr\left\lbrace  Y=1 \right\rbrace + \Pr\left\lbrace \phi(X,U)=1 \right\rbrace -2\Pr\left\lbrace \phi(X,U)=1, Y=1 \right\rbrace $, this performance measure that should be  low decreases as $ \Pr\left\{ \left[\phi\left(X\mbox{,}U\right)=1\right]\wedge\left[Y=1\right]\right\}$ increases. This threshold will be one of the main turning parameters  of the learning procedure that should be set carefully in order to focus on pattern-binary classifier which not performs too weak.

The following definition highlight how a pattern-based binary classifier can be considered as a special case of association rules.
\begin{defn}
A binary association rule is an association rule of the form $U\rightarrow\left[Y=1\right]$
where $U=\left(m_{h_{l}}^{X_{l}}\right)_{l\in L}$ is a pattern based
on marginal components of $X$, $X_{l}\mbox{,}l\in L\subset\left\{ 1\mbox{,}\cdots\mbox{,}p\right\} $. 
\end{defn}
In an ealier machine learning paper devoted to binary association rules that they called classification rules, Liu et al.\cite{liu_integrating_1998} defined the local support of a binary association rule $U\rightarrow Y$ as the conditional probability
 $\Pr\left\{ \left[{\displaystyle \prod_{l\in L}}\left[X_{l}=m_{h_{l}}^{X_{l}}\right]=1\right]\mid\left[Y=1\right]\right\} $.
This definition is anything else only the definition of the sensitivity
of the classifier $\phi\left(X\mbox{,}U\right)$ evoked in the previously.
Moreover one understands easily that the confidence of the binary
association rule $U\rightarrow\left[Y=1\right]$ is equal to the positive
predictive value of the classifier $\phi\left(X\mbox{,}U\right)$. So in the sequel we will focus  on binary association rules as special cases of association rules  that may be suitable for classification task. \\

\section{Strategy for learning classification using binary association rules}
The statistical learning method that we propose in this analysis requires to discretize all continuous variables of the study data. Several methods of discretization of a numerical variable have been proposed in the literature \cite{clarke_entropy_2000}. We suggest that readers use the entropy based method to discretize all numeric variables. This is the method we used in this analysis. The entropy based method allow to choose the partitioning points in a sorted set of continuous values to minimize the joint entropy of the continuous variable and the response variable \cite{fayyad_handling_1992,Fayyad93}. The method is expanded to minimize a MDLP (Minimum Description Length Principle) metric to choose the partitioning points. This is an effective method to improve  the  decision tree learning  and the Bayesian naive classifier for classification by recursively finding more partitioning points\cite{an_discretization_1999}. However the main steps of our proposed statistical learning method can be outlined in the following steps.

\begin{enumerate}\label{chap3_algorithme1}
\item We process a training set by using apriori algorithm for mining all association rules.  Apriori is one of the most widely implemented association rules mining algorithms that pioneered the use of support-based pruning to systematically control the exponentially growth of candidate rules. In the following, we focus on binary association rules generated by the apriori algorithm and that satisfy the following learning conditions: $support$ higher than $s_0$,and $confidence$ higher than $c_0$.
At the end of this step, a large set of patterns $\mathcal{U}_{\lambda}$ is generated. The set of patterns $\mathcal{U}_{\lambda}$ contain both redundant patterns and it is  defined as follow 
\[
\mathcal{U}_{\lambda} = \left\{ U=\left(m_{h(j)}^{X_j} \right)_{j \in J} ; \,  \Pr(Y=1,\phi(X,U)=1) > s_0, \Pr(Y=1|\phi(X,U)=1) > c_0 \right\}
\]
Where $\lambda =(s_0,c_0)$ is the learning parameter of the set $\mathcal{U}_{\lambda}$. In practice, one can extend  the learning parameter $\lambda$ by adding a significance level for Fisher's exact test $t_0$ and a maximum threshold for the size of a pattern $l_0$.\\
It is therefore necessary to prune redundant patterns in order to obtain a reduced set $\mathcal{U}_{\lambda}^{\prime}$ containing only frequent and non-redundant  patterns.

\item From a validation set, we reassessed first all performance indicators (sensitivity, specificity, positive predictive value, etc..). Then we remove all patterns which the positive likelihood ratio is less than one or support is equal to zero. Then throughout the remaining patterns, we look for nested patterns. When we have two patterns are nested we select the patterns that has the most significant positive predictive value. At the end of the step 2 we have a set of patterns $\mathcal{U}_{\lambda}^{''}$ such that $|\mathcal{U}_{\lambda}^{''}| \leq |\mathcal{U}_{\lambda}^{\prime}|$.
\item We define the classification rule (classifier) $ \phi $ as being a function of all the patterns of the set $ \mathcal{U}_{\lambda}^{''}$. Let $x$ be an observation, we have
\[
\phi(X,\lambda)(x)= \left\lbrace \begin{array}{rl}
1 & \textrm{ if } \quad {\displaystyle \sum_{m=1}^{|\mathcal{U}_{\lambda}^{''}|} \phi(X,U_m)(x) > 0} \\ \\
0 & \textrm{ else}
\end{array}
\right. 
\]
We propose to classify  positive an observation $X$ when it verifies at least one pattern among those in the set $\mathcal{U}_{\lambda}^{''}$. This means that an observation $x$ of $X$ is classified as positive if it verifies at least one pattern among those in the set $\mathcal{U}_{\lambda}^{''}$.
\end{enumerate}

In the following, we focus on patterns generated from the apriori algorithm that are both correlated with the response variable and which satisfy the following learning conditions: $support \geq s_0$,\, $confidence \geq c_0$ and \, $size \leq l_0$. At the end of this step, a large set, containing both redundant profiles and profiles with low performance, is generated. Therefore it is  necessary to develop a strategy for pruning redundant patterns in order to obtain a reduced set containing only frequent and non-redundant  patterns.
Moreover we can state that:

\begin{defn} Let $\, U_1=\left( m_{h_l}^{X_l}\right) _{l\in L}\,$ and $\, U_2=\left( m_{h_j}^{X_j}\right) _{j\in J}\,$ be two nested patterns such that $L\subset J$. The pattern $U_k$, $k \in \{1,2\}$, is redundant with respect to $U_{k'}$, $k'\in \{1,2\}$ and $k' \neq k$, if the classification function $\phi(X,U_{k'})$ 
has better performance indicators than the classification function $\phi(X,U_k)$.
\end{defn}
In practice it is not useful to have redundant patterns in a classifier because this will produce an over-fitting classifier. To avoid generating an over-fitting classifier, we will state in the following section how to dealing redundant pattern in order to remove them in final set of patterns that will constitute the classifier.

\section{Redundancy analysis}

\subsection{Some properties of nested patterns useful in redundancy analysis}
Like  standard classification methods, our selection procedure consist to find  patterns that can contribute to improve the classification rule. It is suitable to bring out some basic principles which could help to pruning association rules that generate very weak classification functions. To this end one will pay attention to the subset of rules whose patterns are nested.

\begin{propo}
Let $\, U=\left( m_{h_l}^{X_l}\right) _{l\in L}\,$ and $\, U^{\prime}=\left( m_{h_j}^{X_j}\right) _{j\in J}\,$ be two patterns. If the pattern $U'$ is nested in pattern $U$ then:
\begin{enumerate}
\item $\Pr\left\lbrace \phi(X,U)=1, Y=1 \right\rbrace \geq \Pr\left\lbrace \phi(X,U')=1, Y=1 \right\rbrace$
\item $\Pr\left\lbrace \phi(X,U)=0, Y=0 \right\rbrace \leq \Pr\left\lbrace \phi(X,U')=0, Y=0 \right\rbrace$
\end{enumerate}
\label{chap2_proposition1}
\end{propo}
Therefore the true-positive rate and the true-negative rate are sorted in the opposite way for the classification functions generated by two patterns if one of them is nested in the second one. 
It is worth to notice that in case where the true-negative rates are equal the classification function generated by the pattern with the smallest size is better since its true-positives rate is the highest. In a similar way if the true-positive rates are equal the classification function generated by the pattern with the highest true-negative rate is the best. This provides a criterion that can help to prune the redundant patterns. 

Processing data with an association rules mining algorithm usually produces a large set of association rules within a huge number among them are redundant each others. We have identified in this analysis three propositions  that cn help in redundancy analysis in order to remove redundant patterns.
\begin{propo} Let $\, U=\left( m_{h_l}^{X_l}\right) _{l\in L}\,$ and $\, U^{\prime}=\left( m_{h_j}^{X_j}\right) _{j\in J}\,$ be two nested patterns such that $L\subset J$. Then $\Pr\left\{ \left[\phi(X,U)=1\right]\right\} $=$\Pr\left\{ \left[\phi(X,U')=1\right]\right\} $ if and only if the both following equalities holds:
\begin{enumerate}
\item $\Pr\left\{ \left[\phi(X,U)=1\right],\left[Y=1\right]\right\} $=$\Pr\left\{ \left[\phi(X,U')=1\right],\left[Y=1\right]\right\} $
\item $\Pr\left\{ \left[\phi(X,U)=0\right],\left[Y=0\right]\right\} $=$\Pr\left\{ \left[\phi(X,U')=0\right],\left[Y=0\right]\right\} $
\end{enumerate}
\end{propo}
Since the proposition tells us that the classification function $\phi(X,U)$ et $\phi(X,U')$ have the same performance if and only if they have equal coverages, we should prefer the shortest pattern. 
Therefore we should  perform a statistical hypothesis testing where the null hypothesis \linebreak $\Pr\left(\left[\phi(X,U)=1\right]\right)=\Pr(\left[\phi(X,U')=1\right])$ is considered against its opposite. And we should discard the pattern $U^{\prime}$ if the null hypothesis is not a statistical evidence argument.

\begin{coro}
Let $\, U=\left( m_{h_l}^{X_l}\right) _{l\in L}\,$ and $\, U^{\prime}=\left( m_{h_j}^{X_j}\right) _{j\in J}\,$ be two patterns such that $U'$ is nested in $U$. the following propositions are equivalent:
\begin{enumerate}
\item 
$\Pr\left\lbrace \phi(X,U)=1 \right\rbrace =\Pr\left\lbrace \phi(X,U')=1 \right\rbrace $

\bigskip

\item 
$\left\lbrace \begin{array}{l}
PPV(U,Y)  = PPV(U',Y) \\ \\
\Pr\left\lbrace \phi(X,U)=1,Y=1 \right\rbrace  = \Pr\left\lbrace \phi(X,U')=1,Y=1 \right\rbrace
\end{array}
\right.$

\bigskip

\item 
$\left\lbrace \begin{array}{l}
Err(U,Y) = Err(U',Y)\\ \\
\Pr\left\lbrace \phi(X,U)=1,Y=1 \right\rbrace = \Pr\left\lbrace \phi(X,U')=1,Y=1 \right\rbrace
\end{array}
\right.$
\end{enumerate}
\label{chap2_corollaire1}
\end{coro}
It result from this corollary that performing  a statistical hypothesis testing \linebreak $H_0 : \Pr\left(\left[\phi(X,U)=1\right]\right)=\Pr(\left[\phi(X,U')=1\right])$  agains $H_1 : \Pr\left(\left[\phi(X,U)=1\right]\right) \neq \Pr(\left[\phi(X,U')=1\right])$ is equivalent to perform statistical hypothesis testing from propositions 2. to 7. mentioned in the Corollary. 

\begin{propo}
Let $\,\displaystyle U=\left( m_{h_l}^{X_l}\right) _{l\in L}$ and $\,\displaystyle U'=\left( m_{h_j}^{X_j}\right) _{j\in J}$ be two patterns such that $U'$ is nested in $U$.\\
 If \,$\Pr\left\lbrace [\phi(X,U)=1], [Y=1] \right\rbrace =\Pr\left\lbrace [\phi(X,U')=1], [Y=1] \right\rbrace $ then
\begin{enumerate}
\item $PPV(U,Y) \leq PPV(U',Y)$
\item $NPV(U,Y) \leq NPV(U',Y)$
\item $PLR(U,Y) \leq PLR(U',Y)$
\item $NLR(U,Y) \geq NLR(U',Y)$

\item $Err(U,Y) \geq Err(U',Y)$
\end{enumerate}
\label{proposition3}
\end{propo}

It comes from the statement above (proposition \ref{proposition3})  that in case of equality of the true-positives rates of two different classification functions generated by two nested patterns, not only the sparsest has the smallest positive predictive value but it has also the smallest negative predictive value and the smallest positive likelihood ratio. It has also the highest misclassification rate and the highest negative likelihood ratio. We can take out the pattern with the smallest size since the classification function which it is associated has weak performance indicators. Therefore one can perform a statistical hypothesis testing where the null hypothesis $\Pr(\left[\phi_{U}=1\right]\mbox{,}\left[Y=1\right])=\Pr(\left[\phi_{U'}=1\right]\mbox{,}\left[Y=1\right])$ is considered against its opposite and discard the pattern $U$ if the null hypothesis is accepted for some pattern $U^{\prime}$. 

\begin{propo}
Let $\,\displaystyle U=\left( m_{h_l}^{X_l}\right) _{l\in L}$ and $\,\displaystyle U'=\left( m_{h_j}^{X_j}\right) _{j\in J}$ be two patterns such that $U'$ is nested in $U$.\\
If\, $\Pr\left\lbrace [\phi(X,U)=0], [Y=0] \right\rbrace =\Pr\left\lbrace [\phi(X,U')=0], [Y=0] \right\rbrace $ \,then
\begin{enumerate}
\item $PPV(U,Y) \geq PPV(U',Y)$
\item $NPV(U,Y) \geq NPV(U',Y)$
\item $PLR(U,Y) \geq PLR(U',Y)$
\item $NLR(U,Y) \leq NLR(U',Y)$
\item $Err(U,Y) \leq Err(U',Y)$
\end{enumerate} 
\label{proposition4}
\end{propo}

It comes from this proposition \ref{proposition4} that if one has two nested patterns such that respective classification functions that are generated by nested patterns have the same true-negative rate  then the classification function generated by the shortest pattern has the highest positive predictive value, the highest negative predictive value and the highest positive likelihood ratio. And it has also the smallest negative likelihood ratio and the smallest misclassification rate. This property has been pointed out first in Jiuyong Li \& al. as the anti-monotonic property \cite{JLiFahey2009}.
We can perform a statistical hypothesis testing where the null hypothesis $\Pr\left(\left[\phi_{U'}=1\right]\mbox{,}\,\left[Y=1\right]^{c}\right)=\Pr\left(\left[\phi_{U}=1\right]\mbox{,}\,\left[Y=1\right]^{c}\right)$ is considered against its opposite and discard the pattern $U^{\prime}$ and all the patterns generated by U (containing U) that are nested in $U^{\prime}$  if the null hypothesis is accepted for some $U$.

\subsection{Dealing with redundancy by using statistical hypothesis testing}
The application of the theoretical results presented in the previous section requires to make a hypothesis test on the equality of the coverages, on the equality of the supports and on the equality of the  specificities of two nested profiles. To achieve this, we propose to use a stochastic test. \\
In principle, if the equalities are not true on the learning sample then we can say it is not on the study population. However, we can not say the same when the equalities are true about the learning sample. This is why a stochastic test (randomized test) is required.

Let $\,\displaystyle U_1=\left( m_{h_l}^{X_l}\right) _{l\in L}$ and $\,\displaystyle U_2=\left( m_{h_j}^{X_j}\right) _{j\in J}$ be two patterns such that $U_2$ is nested in $U_1$. Let $\phi(X,U_1)$ and $ \phi(X,U_2)$ be two classification functions generated by $U_1$ and $U_{2}$ respectively.  

\begin{enumerate}
\item[\textbf{(a)}] When trying to test the equality of coverages of two nested patterns, we can consider the parameter $\theta_1$ defined by $\theta_1=\Pr(\phi(X,U_1)=1 )-\Pr(\phi(X,U_2)=1)$. We want to decide whether or not $\theta_1$ is zero e.g to decide  between two hypotheses : 
$ H^1_0 : \theta_1 = 0 \quad vs \quad H^1_1 : \theta_1 \neq 0 $.\,
We will consider the random variable defined by  $ Z_1(X)=\phi(X,U_1)-\phi(X,U_2)$. 
Given that $U_2$ is nested in $U_1$, we have  $[\phi(X,U_2)=1]\subset [\phi(X,U_1)=1]$. And then we have 
\[
Z_1(X)=\left\{
\begin{array}{rl}
1 & \textrm{ si } \phi(X,U_1)=1 \textrm{ et } \phi(X,U_2)=0\\ \\
0 & \textrm{ si } \phi(X,U_1)=\phi(X,U_2)
\end{array}
\right.
\]
 
\bigskip
\item[\textbf{(b)}] When trying to test the equality of supports of two nested patterns, we can consider the parameter $\theta_2$ defined by  $\theta_2=\Pr([\phi(X,U_1)=1,Y=1] )-\Pr([\phi(X,U_2)=1,Y=1])$. We want to decide between two hypotheses : 
$ H^2_0 : \theta_2 = 0 \quad vs \quad H^2_1 : \theta_2 \neq 0 $.\,
We will associated to this hypotheses test the random variable $Z_2(X)$ defined by
\[
Z_2(X)= \ind_{}\left([\phi(X,U_1)=1,Y=1]\right) - \ind_{}\left([\phi(X,U_2)=1,Y=1]\right)
\]
Given that $U_2$ is nested in $U_1$, we have  $[\phi(X,U_2)=1,Y=1]\subset [\phi(X,U_1)=1,Y=1]$. And then we can write that 
\[
Z_2(X)=\left\{
\begin{array}{rl}
1 & \textrm{ si } \ind_{}\left(\phi(X,U_1)=1,Y=1]\right)=1 \textrm{ et } \ind_{}\left(\phi(X,U_2)=1,Y=1]\right)=0\\ \\
0 & \textrm{ si } \ind_{}\left(\phi(X,U_1)=1,Y=1]\right)= \ind_{}\left([\phi(X,U_2)=1,Y=1]\right)
\end{array}
\right.
\]

\bigskip
\item[\textbf{(c)}] When trying to test the equality of specificities (false negative rates) of two nested profiles, we can consider the parameter $\theta_3$ defined by  $\theta_3=\Pr([\phi(X,U_2)=0,Y=0] )-\Pr([\phi(X,U_1)=0,Y=0])$. We want to decide between two hypotheses :  
$ H^3_0 : \theta_3 = 0 \quad vs \quad H^3_1 : \theta_3 \neq 0 $. \,
The random variable $Z_3(X)$ associated to the hypotheses test is defined by
\[
 Z_3(X)=\ind_{}\left([\phi(X,U_2)=0,Y=0]\right)-\ind_{}\left([\phi(X,U_1)=0,Y=0]\right)
\]
Given that $U_2$ is nested in $U_1$, we have  $[\phi(X,U_2)=0,Y=0]\supset [\phi(X,U_1)=0,Y=0]$, So we can write that  
\[
Z_3(X)=\left\{
\begin{array}{rl}
1 & \textrm{ si } \ind_{}\left([\phi(X,U_1)=0,Y=0]\right)=0 \textrm{ et } \ind_{}\left([\phi(X,U_2)=0,Y=0]\right)=1\\ \\
0 & \textrm{ si } \ind_{}\left([\phi(X,U_1)=0,Y=0]\right)=\ind_{}\left([\phi(X,U_2)=0,Y=0]\right)
\end{array}
\right.
\]
\end{enumerate} 
The random variables $(Z_k(X))_{k=1:3}$ are Bernoulli variables with parameters $(\theta_k)_{k=1:3}$ respectively.

Let $\mathcal{D}_n=(X_i,Y_i)_{i \in 1:n}$ be a sample set of $n$ observations from the pair of random variables $(X,Y)$. Given that the observations $(X_i)_{i=1:n}$ are independent then $Z_k(X_i)_{i=1:n}$ are independent realisations. We can deduce that $\sum_{i=1}^n Z_k(X_i)$ is a realisation of a random variable following a binomial distribution $ \mathcal{BN}(n,\theta_k)$. For all $k \in \{1:3\}$, the statistical test $\varphi_k\left(\mathcal{D}_n\right)$ is defined by:
\[
\varphi_k\left(\mathcal{D}_n\right)=\left\{ \begin{array}{ll}
1 & \textrm{ si } \sum_{i=1}^n Z_k(X_i) > 0\\ \\
1-\gamma_k & \textrm{ si } \sum_{i=1}^n Z_k(X_i) = 0 \quad et \quad  0 < \gamma_k \leq 1
\end{array}
\right.
\]
We take a number $\mu$ uniformly distributed between $0$ and $1$. if $\mu \geq 1-\gamma_k$ we reject $H^k_0$ and if $\mu < 1-\gamma_k$ we accept $H^k_0$ with $0 < \gamma_k \leq 1$. The stochastic test is used as follow :
\begin{itemize}
\item If $\varphi_k(\mathcal{D}_n)=1$ : reject  $H^k_0$
\item If $\varphi_k(\mathcal{D}_n)=1-\gamma_k$ : reject  $H^k_0$ with probability $\gamma_k$  
\end{itemize}

The test level is obtained from :
\begin{eqnarray*}
\Pr\left(\textrm{reject } H^k_0|H^k_0\right) &=&\Pr\left(\varphi_k(\mathcal{D}_n)=1\,|\,H^k_0\right)+\Pr\left(\varphi_k(\mathcal{D}_n)=1-\gamma_k, \mu \geq 1-\gamma_k\,|\,H^k_0\right)\\
&=& \Pr\left(\sum_{i=1}^n Z_k(X_i) > 0\,|\,H^k_0\right) + \Pr\left(\sum_{i=1}^n Z_k(X_i) = 0\,|\,H^k_0\right)\Pr\left(\mu \geq 1- \gamma_k\right)\\
&=& \gamma_k \qquad \textrm{given that} \quad \Pr\left(\sum_{i=1}^n Z_k(X_i) = 0\,|\,H^k_0\right)=1
\end{eqnarray*} 

And the test power is got from 
\begin{eqnarray*}
\Pr\left(\textrm{reject } H^k_0|H^k_1\right)  &=& \Pr\left(\varphi_k(\mathcal{D}_n)=1\,|\,H^k_1\right)+\Pr\left(\varphi_k(\mathcal{D}_n)=1-\gamma_k, \mu \geq 1-\gamma_k\,|\,H^k_1\right)\\
&=& 1-\Pr\left(\sum_{i=1}^n Z_k(X_i) = 0\,|\,H^k_1\right)\Pr\left(\mu < 1- \gamma_k\right)\\
&=& 1-(1-\theta_k)^n(1-\gamma_k)
\end{eqnarray*}

The stochastic test presented above can be used regardless of the size of the training set. However the test becomes more powerful when the data size becomes larger.\\

To end this section, we summarize all of the different steps in an algorithm that we present below. This algorithm summarize the hypothesis testing on equal coverages  of two nested patterns,  the hypothesis testing on equal true-positive rates  of two nested patterns step and the hypothesis testing on equal false-positive rates  of two nested patterns step.


\scriptsize
\begin{algorithm}[H]
\caption{Pruning procedure of the redundant patterns}
\begin{algorithmic}
\REQUIRE $\mathcal{U}_{\lambda}$ a set of frequent patterns
\ENSURE $\mathcal{U}^{\prime}_{\lambda}$ a set of frequent and non-redundant patterns\\~~\\

\FORALL{ $U \in \mathcal{U}_{\lambda} $}
\STATE $\mathcal{S}_{U} \leftarrow subset(U,\mathcal{R})$
\FORALL{$U^{\prime} \in \mathcal{S}_{U}$}
\STATE Test the following hypotheses:
\STATE $ H^1_0: \Pr\left\{ \phi(X,U)=1\right\} = \Pr\left\{ \phi(X,U')=1\right\}$ vs $ H^1_1: \Pr\left\{ \phi(X,U)=1\right\} \neq \Pr\left\{ \phi(X,U')=1\right\}$
\IF{$H^1_0$ is true}
\STATE $\mathcal{S'}_{U} \leftarrow delete(U',\mathcal{S}_{U})$
\ELSE 
\STATE Test the following hypotheses:
\STATE $ H^2_0: \,\Pr\left\{ \phi(X,U)=1\mbox{,}Y=1\right\} = \Pr\left\{ \phi(X,U')=1\mbox{,}Y=1\right\}$ vs $ H^2_1: \,\Pr\left\{ \phi(X,U)=1\mbox{,}Y=1\right\} \neq \Pr\left\{ \phi(X,U')=1\mbox{,}Y=1\right\}$ 
\IF{$H^2_0$ is true}
\STATE $\mathcal{S'}_{U} \leftarrow delete(U,\mathcal{S}_{U})$ 
\ELSE
\STATE Test the following hypotheses:
\STATE $ H^3_0: \,\Pr\left\{ \phi(X,U)=0\mbox{,}Y=0\right\} = \Pr\left\{ \phi(X,U')=0\mbox{,}Y=0\right\}$ vs $ H^3_1: \,\Pr\left\{ \phi(X,U)=0\mbox{,}Y=0\right\} \neq \Pr\left\{ \phi(X,U')=0\mbox{,}Y=0\right\}$
\IF{$ H^3_0$ is true}
\STATE  $\mathcal{S'}_{U} \leftarrow delete(U',\mathcal{S}_{U})$
\ENDIF
\ENDIF
\ENDIF
\ENDFOR
\ENDFOR
\STATE $\mathcal{U}^{\prime}_{\lambda} \leftarrow \bigcup_{U\in \mathcal{U}_{\lambda}} \mathcal{S'}_{U}$
\end{algorithmic}
\end{algorithm}
\normalsize

Generally the set $\mathcal{U}^{\prime}_{\lambda} $ contains a large number of patterns whose majority is not relevant to construct a classification function effective and easy to implement. To remove less relevant patterns, we propose to use a pruning procedure based on the positive predictive value.
\section{Selecting a set of relevant patterns}

After the pruning step of the redundant patterns, we obtain a reduced set of patterns. We note that the redundant patterns pruning procedure does not eliminate all nested patterns. The selecting procedure of the  relevant patterns aim to compare the nested remaining patterns and select the most relevant. In summary, a test is used to compare the positive predictive values of the nested patterns. 

\subsection{Selecting a set of relevant patterns when sample is with large size}
In general, we can use a comparison test positive predictive values of two nested profles to select the most appropriate. This test is based on the asymptotic normality of the logarithm of the ratio of the positive predictive values of nested patterns.
\begin{propo}
Let ${\displaystyle U_1=\left( m_{h_j}^{X_j}\right) _{j\in J}}$ and ${\displaystyle U_2=\left( m_{h_l}^{X_l}\right) _{l\in L}}$ be two patterns such that $U_2$ is nested in $U_1$. Let $\widehat{PPV}(U_1,Y)$ and $\widehat{PPV}(U_2,Y)$ be the empirical estimators of $PPV(U_1,Y)$ and  $PPV(U_2,Y)$ respectively. The random variable ${\displaystyle log\left( \frac{\widehat{PPV}(U_1,Y)}{\widehat{PPV}(U_2,Y)}\right) }$ is asymptotically distributed according to a centred normal distribution  with  a  variance $\Sigma$  defined by: 
\begin{eqnarray*}
 \Sigma &=& \sum_{i=1}^6p_i \nabla_i^2 -\left( \sum_{i=1}^6p_i\nabla_i\right)^2 
\end{eqnarray*}
where 
\[
\left( 
\begin{array}{c}
\nabla_1\\ \nabla_2\\ \nabla_3\\ \nabla_4\\ \nabla_5\\ \nabla_6
\end{array}
\right)  = \left( 
\begin{array}{c}
\frac{1}{p_1+p_4} +  \frac{1}{p_1+p_2} - \frac{1}{p_1} - \frac{1}{p_1+p_2+p_4+p_5}\\
\frac{1}{p_1+p_2} - \frac{1}{p_1+p_2+p_4+p_5}\\
0\\ 
\frac{1}{p_1+p_4} - \frac{1}{p_1+p_2+p_4+p_5}\\
-\frac{1}{p_1+p_2+p_4+p_5}\\
0
\end{array}
\right) 
\]
$ p_1$, $p_2$, $p_3$, $p_4$, $p_5$ and $p_6$ are respective probabilities of the events 
\begin{eqnarray*}
E_1 = \{Y=1,\phi(X,U_1)=1,\phi(X,U_2)=1\} & & E_2 = \{Y=1,\phi(X,U_1)=1,\phi(X,U_2)=0\} \\
E_3 = \{Y=1,\phi(X,U_1)=0,\phi(X,U_2)=0\} & & E_4 = \{Y=0,\phi(X,U_1)=1,\phi(X,U_2)=1\} \\
E_5 = \{Y=0,\phi(X,U_1)=1,\phi(X,U_2)=0\} & & E_6 = \{Y=0,\phi(X,U_1)=0,\phi(X,U_2)=0\}
\end{eqnarray*}
such that $\sum\limits_{i=1}^6 p_i =1$
\end{propo}
Given that the pattern $U_2$ is nested in the pattern $U_1$, we have
\begin{eqnarray*}
PPV(U_1,Y) &=& \frac{\Pr\{Y=1,\phi(X,U_1)=1\}}{\Pr\{\phi(X,U_1)=1\}} = \frac{p_1+p_2}{p_1+p_2+p_4+p_5}\\~~\\
PPV(U_2,Y) &=& \frac{\Pr\{Y=1,\phi(X,U_2)=1\}}{\Pr\{\phi(X,U_2)=1\}} = \frac{p_1}{p_1+p_4}
\end{eqnarray*}
To obtain the asymptotic normality of the logarithm of the ratio of $PPV(U_1,Y)$ and $PPV(U_2,Y)$, we consider  the variable $Z= (I_{E_1}, \dots, I_{E_6})$ distributed according to a generalized Bernoulli distribution  of parameter $\theta = (p_1,\dots,p_6)$ whose covariance matrix is given by 
\[
\Lambda(\theta) = diag(\theta) - \theta^{T}\theta
\]
where $I_{E_k}$, $k=1:6$ is the indicator function of the event $E_k$.

Let $(Z_i)_{i=1:n}$ be a sequence of $n$ random variables identically distributed according to a generalized Bernoulli distribution where $n$ is supposed to be large enough. Let denote $\widehat{\theta}_n$ the empirical estimator of the parameter $\theta$. It is defined by 
\[
{\displaystyle \widehat{\theta}_n = \frac{1}{n}\sum_{i=1}^{n} Z_i }
\]
And let denote by $g(\theta)$ be the logarithm of the ratio of $PPV(U_1,Y)$ and $PPV(U_2,Y)$ defined by
\[
g(\theta) = log\left( \frac{PPV(U_1,Y)}{PPV(U_2,Y)} \right) =  log\left( \frac{(p_1+p_4)(p_1+p_2)}{p_1(p_1+p_2+p_4+p_5)} \right)
\]
It result from the central limit theorem that 
\[
\sqrt{n}\left( \widehat{\theta}_n - \theta \right) \xrightarrow[]{\ \mathcal{L}\ } \mathcal{N}\left( 0,\Lambda(\theta) \right) 
\]
Using the multivariate delta method, we obtain that
\[
\sqrt{n}\left( g(\widehat{\theta}_n) - g(\theta) \right) \xrightarrow[]{\ \mathcal{L}\ } \mathcal{N}\left( 0,{}^T\nabla g(\theta)\Lambda(\theta)\nabla g(\theta) \right) 
\]
where
\[
\nabla g(\theta) = \left( 
\begin{array}{c}
\nabla_1\\ \vdots \\ \nabla_6
\end{array}
\right) = \left( 
\begin{array}{c}
\frac{1}{p_1+p_4} +  \frac{1}{p_1+p_2} - \frac{1}{p_1} - \frac{1}{p_1+p_2+p_4+p_5}\\
\frac{1}{p_1+p_2} - \frac{1}{p_1+p_2+p_4+p_5}\\
0\\ 
\frac{1}{p_1+p_4} - \frac{1}{p_1+p_2+p_4+p_5}\\
-\frac{1}{p_1+p_2+p_4+p_5}\\
0
\end{array}
\right) 
\]
and 
\begin{eqnarray*}
{}^T\nabla g(\theta)\Lambda(\theta)\nabla g(\theta) 
&=& \sum_{i=1}^6p_i \nabla_i^2 -\left( \sum_{i=1}^6p_i\nabla_i\right)^2 
\end{eqnarray*}
Using  the continuity of the $\theta \longmapsto \nabla g(\theta) $ and $ \theta \longmapsto \Lambda(\theta)$ applications and the almost sure convergence of $\widehat{\theta}_n$ to $\theta$, we show that
\[
{}^T\nabla g(\widehat{\theta}_n)\Lambda(\widehat{\theta}_n)\nabla g(\widehat{\theta}_n) \xrightarrow[]{\ p.s\ } {}^T\nabla g(\theta)\Lambda(\theta)\nabla g(\theta)
\] 
It result from the Slutsky theorem that 
\[
\frac{\sqrt{n}\left(g(\widehat{\theta}_n) - g(\theta) \right) }{\sqrt{{}^T\nabla g(\widehat{\theta}_n)\Lambda(\widehat{\theta}_n)\nabla g(\widehat{\theta}_n)}} \xrightarrow[]{\ \mathcal{L}\ } \mathcal{N}(0,1)
\] 
Under the null hypothesis, we have $g(\theta) = 0$. This allows us to build the following strategy to select the most relevant patterns.

\begin{enumerate}
\item Select the pattern  $U_2$ if 
\[
g(\widehat{\theta}_n) < -q_{1-\alpha/2}\sqrt{\frac{{}^T\nabla g(\widehat{\theta}_n)\Lambda(\widehat{\theta}_n)\nabla g(\widehat{\theta}_n)}{n}}
\]
\item Select the pattern $U_1$  if
\[
g(\widehat{\theta}_n) \geq -q_{1-\alpha/2}\sqrt{\frac{{}^T\nabla g(\widehat{\theta}_n)\Lambda(\widehat{\theta}_n)\nabla g(\widehat{\theta}_n)}{n}}
\]
\end{enumerate}
where $q_{1-\alpha/2}$ is the quantile of order $1-\alpha/2$ of the standard normal distribution.

\begin{algorithm}[H]
\caption{Selecting procedure of the relevant patterns}
\begin{algorithmic}
\REQUIRE $\mathcal{D}$ a validation data set; $\mathcal{U}_{\lambda}$ a set of frequent and non redundant patterns
\ENSURE  $\mathcal{U}^{\prime}_{\lambda}$ a set of relevant patterns \\~~\\

\FORALL{ $C \in \mathcal{U}_{\lambda}$}
\STATE $S \leftarrow subset (C,\mathcal{U}_{\lambda} )$
\FORALL{$C^{\prime} \in S $}
\STATE $\widehat{\theta}_n \leftarrow (p_1,\dots,p_6|\mathcal{D})$
\STATE $g(\widehat{\theta}_n) \leftarrow log(VPP(C,Y|\widehat{\theta}_n)) - log(VPP(C',Y|\widehat{\theta}_n))$
\STATE $ \Lambda(\widehat{\theta}_n) \leftarrow diag(\widehat{\theta}_n) - \widehat{\theta}_n^t\widehat{\theta}_n$
\STATE $\nabla_n \leftarrow \nabla g(\widehat{\theta}_n)$
\ENDFOR
\IF{ there is $C^{\prime} \in S$ such that ${\displaystyle g(\widehat{\theta}_n) < -q_{1-\alpha/2}\sqrt{\frac{\nabla_n^t\Lambda(\widehat{\theta}_n)\nabla_n}{n}}}$}
\STATE $ \mathcal{U}^{\prime}_{\lambda} \leftarrow delete(C,\mathcal{U}_{\lambda}) $
\ELSE
\STATE $ \mathcal{U}^{\prime}_{\lambda} \leftarrow delete(S,\mathcal{U}_{\lambda}) $
\ENDIF
\ENDFOR
\end{algorithmic}
\end{algorithm}

The learning method, as described previously, requires a large database that will be subdivided into three subsets of sufficiently large sizes (learning, validation and test). In the task of machine learning, it is common to encounter data whose size does not allow a subdivision of observations. Faced with such data, we can consider a bootstrap procedure.

\subsection{Selecting a set of relevant patterns when sample is with small size}
According to the central limit theorem, the following condition is true only when the number of observations is large enough. 
\[
S_{n}=\frac{\sqrt{n}\left(g(\widehat{\theta}_n) - g(\theta) \right) }{\sqrt{{}^T\nabla g(\widehat{\theta}_n)\Lambda(\widehat{\theta}_n)\nabla g(\widehat{\theta}_n)}} \xrightarrow[]{\ \mathcal{L}\ } \mathcal{N}(0,1) 
\] 
where $n$ is the observations size.

In the case where the number of observations is small, it is not possible to have this condition for selecting a set of relevant patterns. This alternative method is to use a bootstrap hypothesis testing. The bootstrap is a well known re-sampling technique. The principle of the bootstrap method is to replace the unknown distribution $ F $ that generated the sample by the distribution $F_{n} $ which associates to each observation a weight $ 1 / n $. So when drawing randomly with replacement $ n $ elements from $ n $ initial observations, we obtain a bootstrap sample of size $ n $ by the empirical distribution $ F_{n}$ \cite{efron_jackknife_1982,efron_introduction_1994}.

Let $g(\theta)$ be our statistic of interest and $F_{g(\theta)}$ its sampling distribution. We can notice that $F_{g(\theta)}$ depend on the generalized Bernoulli distribution $G_Z$ of the random variable $Z$ whose observed values are $z_1, \dots, z_n$. Someone can write $F_{g(\theta)} = F_{g(\theta)}(\cdot,G_Z)$, where $G_Z$ depend on the distribution $F_X$ of the random variable $X$ whose observed values are $x_1, \dots, x_n$. In summary the sampling distribution $F_S$ depend on the realisations $z$ of the random variable $Z$ and the distribution $F_X$.  We note 
\[
F_{g(\theta)}= F_{g(\theta)}(\cdot,z,F_X)
\]
Since the $F_X$ distribution is unknown, we can replace it by the empirical distribution $F_n$ of the observed values $x_1, \dots, x_n$ in the previous equality. Replace the unknown distribution $F_X$ by the empirical distribution $F_n$ is equivalent to randomly draw with replacement $n$ elements   from the original data $x_1, \dots, x_n$.\\
Let $g(\widehat{\theta}_n)$, a function of the sample $X_1, \dots , X_n$, denote an estimator of the unknown quantity $g(\theta)$, and write $g(\widehat{\theta}^{*}_{n})$ the value of $g(\widehat{\theta}_n)$ computed from a bootstrap sample $X^{*}_{1}, \dots, X^{*}_{n}$ drawn from the original sample with replacement. We denote $\widehat{\sigma}_n = \sqrt{ \frac{1}{n} \left( {}^T\nabla g(\widehat{\theta}_n)\Lambda(\widehat{\theta}_n)\nabla g(\widehat{\theta}_n) \right)} $ the standard deviation of  $g(\widehat{\theta}_n)$. Let $\widehat{\sigma}^{*}_{n}$ denote the value of $\widehat{\sigma}_n$ computed for the bootstrap sample rather than the sample. Then  the bootstrap distribution of $\left( g(\widehat{\theta}^{*}_{n}) - g(\widehat{\theta}_{n}) \right) \big/\widehat{\sigma}^{*}_{n}$ estimates the distribution of $\left(g(\widehat{\theta}_{n}) - g(\theta) \right)\big/\widehat{\sigma}_{n}$ under the null hypothesis \cite{hall_two_1991}. To make hypotheses test with a null hypotheses $ H_0 : g(\theta) = 0 $ a gains a n alternative hypotheses $H_1 : g(\theta) \neq 0$, we proceed as follow :
\begin{itemize}
\item First, we compute the value $s^{0}_{n}$ of the statistic $S_{n}$ for the sample $X_1, \dots , X_n$.
\item Second, we simulate $B$ resamples $X^{b}_{1}, \dots, X^{b}_{n}$ \, $(b = 1,\dots, B)$  drawn from the sample with replacement. For each resample, we denote $s^{b}_{n}$ the value of $S_{n}$ computed the $b^{th}$ resample. 
\[
s^{b}_{n} = \frac{g(\widehat{\theta}^{b}_{n}) - g(\widehat{\theta}_{n})}{\widehat{\sigma}^{b}_{n}}
\]
\item third, we compute the bootstrap $p-value$
\[
p_{n} = \frac{1}{B} \sum_{b=1}^{B}I\left(S^{b}_{n} > s^{0}_{n} \right)
\]
\end{itemize}
This allows us to build the following strategy to select the most relevant patterns.
\begin{enumerate}
\item[(a)] Select the pattern $U_2$ if $p_n < \alpha/2$
\item[(b)] Select the pattern $U_2$ if $p_n \geq \alpha/2$
\end{enumerate}
where $\alpha$ is the level of the test. We can notice that this hypothesis test allow to favour the shorter patterns.

\begin{algorithm}[H]
\caption{Selecting procedure of the relevant patterns}
\begin{algorithmic}
\REQUIRE $\mathcal{D}$ a validation data set ; $\mathcal{U}_{\lambda}$ a set of frequent and non redundant patterns ; $\alpha$ the level of the test and $B$ the number of bootstrap samples.
\ENSURE  $\mathcal{U}^{\prime}_{\lambda}$ a set of relevant patterns \\~~\\

\FORALL{ $C \in \mathcal{U}_{\lambda}$}
\STATE $S \leftarrow subset (C,\mathcal{U}_{\lambda} )$
\FORALL{$C^{\prime} \in S $}
\STATE $\widehat{\theta}_n \leftarrow (p_1,\dots,p_6|\mathcal{D})$
\STATE $g(\widehat{\theta}_n) \leftarrow log(VPP(C,Y|\widehat{\theta}_n)) - log(VPP(C',Y|\widehat{\theta}_n))$
\STATE $ \Lambda(\widehat{\theta}_n) \leftarrow diag(\widehat{\theta}_n) - \widehat{\theta}_n^t\widehat{\theta}_n$
\STATE $\nabla_n \leftarrow \nabla g(\widehat{\theta}_n)$
\STATE $\widehat{\sigma}_n \leftarrow \sqrt{\frac{1}{n} \left( \nabla^t_n \Lambda \left( \widehat{\theta}_n\right) \nabla_n\right)}$
\STATE $s^0_n \leftarrow g(\widehat{\theta}_n)/ \widehat{\sigma}_n$

\FORALL{ bootstrap sample $\mathcal{D}^{b}$}
\STATE $\widehat{\theta}^{b}_n \leftarrow (p_1,\dots,p_6|\mathcal{D}^{b})$
\STATE $g(\widehat{\theta}^{b}_n) \leftarrow log(VPP(C,Y|\widehat{\theta}^{b}_n)) - log(VPP(C',Y|\widehat{\theta}^{b}_n))$
\STATE $ \Lambda(\widehat{\theta}^{b}_n) \leftarrow diag(\widehat{\theta}^{b}_n) - (\widehat{\theta}^{b}_n)^t\widehat{\theta}_n$
\STATE $\nabla_n \leftarrow \nabla g(\widehat{\theta}^{b}_n)$
\STATE $\widehat{\sigma}^{b}_n \leftarrow \sqrt{\frac{1}{n} \left( \nabla^t_n \Lambda \left( \widehat{\theta}^{b}_n\right) \nabla_n\right)}$
\STATE $s^b_n \leftarrow \left( g(\widehat{\theta}^{b}_n) - g(\widehat{\theta}_n) \right)/ \widehat{\sigma}^{b}_n$
\ENDFOR
\STATE \qquad \qquad $p_{n} \leftarrow \frac{1}{B} \sum_{b=1}^{B}I\left(S^{b}_{n} > s^{0}_{n} \right)$
\STATE
\IF{$p_{n} < \alpha/2$}
\STATE
\STATE \qquad \qquad $\mathcal{U}^{\prime}_{\lambda} \leftarrow delete(C, \mathcal{U}_{\lambda})$
\ELSE 
\STATE \qquad \qquad $\mathcal{U}^{\prime}_{\lambda} \leftarrow delete(C^{\prime}, \mathcal{U}_{\lambda})$
\ENDIF
\ENDFOR
\ENDFOR
\end{algorithmic}
\end{algorithm}

\section{Empirical study}
This section is about  to evaluate our statistical learning method that we denote by \textbf{PBBC} (\textbf{P}attern-\textbf{B}inary \textbf{B}ased \textbf{C}lassifier) on literature data  and to compare its performances to standard classification methods. All the literature data that we have used for evaluating the PBBC are coming from UCI machine learning repository data sets \cite{Bache_machine_2013}.   All of them have two classes. The analysis of the proposed methods were performed in the  R  environment for statistical computing \cite{R_envir_2013}. The Association rules were explored by using the arules \cite{agrawal_fast_1994} package  in the  R  environment for statistical computing. The Table \ref{table1} is shows the dataset, the size in number of observations, the nominal and numerical attributes and the percentage of observations of the minority class.

\begin{table}[H]
\begin{center}
\footnotesize
\begin{tabular}{ccccc}
\hline\\
DATASET & SIZE & \multicolumn{2}{c}{ATTRIBUTES} & $\%$ TARGET CLASS \\
\cline{3-4}
        &       & Nominal     & Numerical &   \\
\hline
Adult                               & 45222 &  8  & 5 & 24.78 \\
Breast Cancer Wisconsin (Original)  & 699   &  10 & 0 & 34.50 \\
Pima Indians Diabetes               & 768   &  0  & 8 & 34.89 \\
\hline
\end{tabular}
\normalsize
\end{center}
\caption{ \footnotesize Datasets used for the evaluations}
\label{table1}
\end{table}

Since the data we have are not very unbalanced, we conducted several experiments by sup-sampling or sub-sampling the databases to obtain unbalanced samples for assessing the statistical learning method. We proceed as follows: we start by selecting the $ n $ observations of the prevailing class and we choose a proportion $ \alpha $ of the rare class. Then we randomly select $ n^{\prime} = n\alpha/(1-\alpha) $ observations of the rare class. Thus, we obtain a sample of $ n + n^{\prime} $ observations which the proportion of the rare class observations is equal to $ \alpha $.

For each simulated sample, we perform many  combinations of the learning parameters  $c_0, s_0, l_0$ (i.e. we propose a dozen combinations of parameters values $\lambda$). Thus each combination produces a classifier for which we can compute its performances : sensitivity, specificity, error, etc. Then we select the classifier that realizes the best performances proceeding using a ROC curve. 

The binary classification from logistic regression or binary regression trees involves fitting a parametric or non-parametric model on the data $\mathcal{D}$. This leads to the evaluation of conditional probabilities $\Pr (Y = 1 | X = x)$ depending on the data $\mathcal{D}$. We obtain a classifier $\phi$ defined by
\[
\phi(x|\alpha) = \left\lbrace
\begin{array}{lll}
1 & \textrm{si} &\Pr(Y=1|X=x,\mathcal{D}) > \alpha \\
0 & \textrm{sinon}&
\end{array}
\right.
\]
where $\alpha \in ]0,1[$\\
In the case of discriminant analysis or Bayesian networks analysis (eg naive Bayesian network), we  consider a prior law  that we denote by $\pi$ for the probability distribution of classes. Then a parametric or non parametric model based on the conditional probabilities  $\Pr(X= x | Y=1, D) $ is adjusted on the data. The obtained classifier $\phi$ is defined as 
\[
\phi(x|\alpha) = \left\lbrace
\begin{array}{lll}
1 & \textrm{si} &\Pr(X=x|Y=1,\mathcal{D})\pi(y) > \alpha \\
0 & \textrm{sinon}&
\end{array}
\right.
\]  
where $\alpha \in ]0,1[$\\
This raises the issue of selecting an optimal classifier based on a compromise on performance measures such as sensitivity, specificity, error rate, etc.. The ROC curve and the AUC measure are generally used to achieve this goal.\\
This approach can be extended to classifier aggregation methods such as binary tree boosting or random forest. Usually these methods use a threshold $ \alpha  = 0.5$ by default. Very often the classifier  $\phi(x|\alpha)$ associated with threshold $\alpha = 0.5$ does not provide better performance. And to compare our method of classification with these methods, we consider the following strategy:
\begin{enumerate}
\item The first step is to determine the optimal learning parameters for adjusting an efficient model. 
\item The second step is identify the optimal probability threshold. In other words, it is to identify the threshold that produces the classifier whose performance measures provides the best compromise.
\item Then, we compare the performance of classifiers obtained with the performance of our classifier. 
\end{enumerate}
This approach has been compared to some competitor methods designed to deal with imbalanced classification problem : SMOTE and ROSE.\\ 
SMOTE is an over-sampling approach in which the minority class is over-sampled by creating "synthetic" examples rather than by over-sampling with replacement. The minority class is over-sampled by taking each minority class sample and introducing synthetic example along the line segments joining any/all the $k$ minority class nearest neighbours. The application of SMOTE  has been performed by choosing $5$ nearest neighbours, as it was suggested by the authors in their paper \cite{Chawla02smote:synthetic}.\\ While ROSE is an an approach based on the generation of new artificial data from the classes, according to a smoothed bootstrap approach. It combines technique of over-sampling and under-sampling by generating an augmented sample of data thus helping the classifier in estimating a more accurate classification rule \cite{menardi_training_2014}. The application of ROSE has been performed by fitting a logistic regression model after to generate an augmented data by ROSE principle. 

The results obtained are shown in Tables below. There were obtained using the \textbf{caret} (\textbf{c}lassification \textbf{a}nd \textbf{re}gression \textbf{t}raining ) package \cite{kuhn_building_2008}. 

\begin{table}[H]
\begin{center}
\footnotesize
\begin{tabular}{lccccccccccc}
\hline\\
 & \multicolumn{5}{c}{\scriptsize $\alpha = 0.007$ } & \multicolumn{6}{c}{\scriptsize $\alpha = 0.015$ }\tabularnewline
{\scriptsize Methods} & {\scriptsize Sensibility}& {\scriptsize Specificity}& {\scriptsize Error}& {\scriptsize AUC}& {\scriptsize PSS}& & {\scriptsize Sensibility}& {\scriptsize Specificity}& {\scriptsize Error}& {\scriptsize AUC}& {\scriptsize PSS}\\
\hline
\hline\\
\textbf{PBBC}&0.815 &0.788 &0.212 &0.801 &0.603 && 0.761 &0.797 &0.204 &0.779 &0.558\\
ROSE      & 0.827 & 0.781 &0.218 &0.804 &0.608 && 0.835 & 0.777 &0.222 &0.806 &0.612\\
SMOTE     & 0.716 & 0.729 &0.271 &0.723 &0.445 && 0.801 & 0.675 &0.323 &0.738 &0.476\\
Random.F  & 0.259 & 0.996 &0.009 &0.628 &0.255 && 0.330 & 0.987 &0.023 &0.658 &0.317\\
Boosting  & 0.210 & 0.999 &0.007 &0.604 &0.208 && 0.278 & 0.996 &0.015 &0.637 &0.275\\
CART      & 0.173 & 0.999 &0.007 &0.586 &0.172 && 0.210 & 0.999 &0.012 &0.605 &0.210\\
CTREE     & 0.185 & 0.999 &0.007 &0.592 &0.184 && 0.205 & 0.999 &0.013 &0.602 &0.203\\
Boost.glm & 0.023 & 0.928 &0.008 &0.525 &-0.049 && 0.026 &0.928 &0.016 &0.523 &-0.045\\
N.Bayes   & 0.160 & 0.999 &0.007 &0.580 &0.160 && 0.131 & 0.999 &0.014 &0.565 &0.130\\ \\

& \multicolumn{5}{c}{\scriptsize $\alpha = 0.03$ } & \multicolumn{6}{c}{\scriptsize $\alpha = 0.07$ }\tabularnewline
{\scriptsize Methods} & {\scriptsize Sensibility}& {\scriptsize Specificity}& {\scriptsize Error}& {\scriptsize AUC}& {\scriptsize PSS}& & {\scriptsize Sensibility}& {\scriptsize Specificity}& {\scriptsize Error}& {\scriptsize AUC}& {\scriptsize PSS}\\
\hline
\hline\\
\textbf{PBBC}&0.773 &0.798 &0.203 &0.786 &0.571 && 0.739 &0.832 &0.175 &0.785 &0.570\\
ROSE       &0.838  &0.787 &0.211 &0.812 &0.625 && 0.811 & 0.777 &0.220 &0.794 &0.589\\
SMOTE      &0.770  &0.727 &0.271 &0.749 &0.498 && 0.791 & 0.778 &0.221 &0.784 &0.569\\
Random.F   &0.406  &0.975 &0.042 &0.691 &0.381 && 0.604 & 0.919 &0.103 &0.762 &0.523\\
Boosting   &0.378  &0.990 &0.028 &0.684 &0.368 && 0.608 & 0.947 &0.076 &0.778 &0.555\\
CART       &0.241  &0.999 &0.024 &0.620 &0.240 && 0.488 & 0.948 &0.084 &0.718 &0.436\\
CTREE      &0.249  &0.997 &0.026 &0.623 &0.246 && 0.568 & 0.940 &0.086 &0.754 &0.508\\
Boost.glm  &0.135  &0.926 &0.030 &0.530 &0.061 && 0.465 & 0.887 &0.080 &0.676 &0.352\\
N.Bayes    &0.143  &0.997 &0.028 &0.570 &0.140 && 0.196 & 0.996 &0.060 &0.596 &0.191\\ \\

& \multicolumn{5}{c}{\scriptsize $\alpha = 0.15$ } & \multicolumn{6}{c}{\scriptsize $\alpha = 0.20$ }\tabularnewline
{\scriptsize Methods} & {\scriptsize Sensibility}& {\scriptsize Specificity}& {\scriptsize Error}& {\scriptsize AUC}& {\scriptsize PSS}& & {\scriptsize Sensibility}& {\scriptsize Specificity}& {\scriptsize Error}& {\scriptsize AUC}& {\scriptsize PSS}\\
\hline
\hline\\
\textbf{PBBC}&0.757 &0.809 &0.199 &0.783 &0.565 && 0.734 &0.832 &0.188 &0.783 &0.566\\
ROSE       &0.836  &0.795 &0.199 &0.815 &0.631 && 0.832  &0.798 &0.196 &0.815 &0.629\\
SMOTE      &0.814  &0.772 &0.222 &0.793 &0.586 && 0.827  &0.776 &0.214 &0.802 &0.603\\
Random.F   &0.790  &0.840 &0.167 &0.815 &0.630 && 0.838  &0.794 &0.198 &0.816 &0.631\\
Boosting   &0.783  &0.870 &0.143 &0.826 &0.653 && 0.819  &0.801 &0.196 &0.810 &0.620\\
CART       &0.503  &0.948 &0.119 &0.725 &0.451 && 0.890  &0.671 &0.285 &0.780 &0.561\\
CTREE      &0.764  &0.857 &0.157 &0.810 &0.621 && 0.833  &0.803 &0.191 &0.818 &0.636\\
Boost.glm  &0.763  &0.761 &0.186 &0.762 &0.525 && 0.854  &0.687 &0.232 &0.771 &0.541\\
N.Bayes    &0.258  &0.991 &0.119 &0.624 &0.249 && 0.293  &0.988 &0.151 &0.640 &0.281\\ \\
\hline
\end{tabular}
\normalsize
\end{center}
\caption{ \footnotesize Sensibility, specificity, error estimation, area under the ROC curve and Pierce score, with different base classification rules, with different proportions ($\alpha$) of the target class, for Adult dataset. This performances are performed on test sample which distribution is identical to distribution of training sample}
\label{table1}
\end{table}

\begin{table}[H]
\begin{center}
\footnotesize
\begin{tabular}{lccccccccccc}
\hline\\
 & \multicolumn{5}{c}{\scriptsize $\alpha = 0.007$ } & \multicolumn{6}{c}{\scriptsize $\alpha = 0.015$ }\tabularnewline
{\scriptsize Methods} & {\scriptsize Sensibility}& {\scriptsize Specificity}& {\scriptsize Error}& {\scriptsize AUC}& {\scriptsize PSS}& & {\scriptsize Sensibility}& {\scriptsize Specificity}& {\scriptsize Error}& {\scriptsize AUC}& {\scriptsize PSS}\\
\hline
\hline\\
\textbf{PBBC}&0.826 &0.678 &0.286 &0.752 &0.503 && 0.839 &0.720 &0.251 &0.780 &0.558\\
ROSE      &0.807 & 0.749 &0.237 &0.778 &0.556 && 0.832 &0.762 &0.221 &0.797 &0.594\\
SMOTE     &0.173 & 0.986 &0.214 &0.580 &0.159 && 0.220 &0.988 &0.201 &0.604 &0.208\\
Random.F  &0.248 & 0.995 &0.189 &0.622 &0.243 && 0.320 &0.987 &0.177 &0.653 &0.307\\
Boosting  &0.232 & 0.999 &0.190 &0.615 &0.231 && 0.270 &0.996 &0.183 &0.633 &0.267\\
CART      &0.169 & 0.999 &0.205 &0.584 &0.168 && 0.150 &0.999 &0.210 &0.575 &0.150\\
CTREE     &0.169 & 0.999 &0.205 &0.584 &0.168 && 0.199 &0.998 &0.199 &0.598 &0.197\\
Boost.glm &0.021 & 0.928 &0.241 &0.526 &-0.051 && 0.022 &0.928 &0.241 &0.525 &-0.050\\
N.Bayes   &0.123 & 0.999 &0.217 &0.561 &0.122 && 0.142 &0.999 &0.212 &0.570 &0.140\\ \\

& \multicolumn{5}{c}{\scriptsize $\alpha = 0.03$ } & \multicolumn{6}{c}{\scriptsize $\alpha = 0.07$ }\tabularnewline
{\scriptsize Methods} & {\scriptsize Sensibility}& {\scriptsize Specificity}& {\scriptsize Error}& {\scriptsize AUC}& {\scriptsize PSS}& & {\scriptsize Sensibility}& {\scriptsize Specificity}& {\scriptsize Error}& {\scriptsize AUC}& {\scriptsize PSS}\\
\hline
\hline\\
\textbf{PBBC}&0.820 &0.755 &0.229 &0.787 &0.575 && 0.788 &0.781 &0.218 &0.784 &0.569\\
ROSE       &0.846  &0.763 &0.217 &0.805 &0.609 && 0.869 &0.751 &0.220 &0.810 &0.620\\
SMOTE      &0.333  &0.979 &0.180 &0.656 &0.312 && 0.415 &0.964 &0.171 &0.689 &0.379\\
Random.F   &0.392  &0.985 &0.161 &0.689 &0.377 && 0.602 &0.949 &0.136 &0.776 &0.551\\
Boosting   &0.269  &0.999 &0.181 &0.634 &0.268 && 0.339 &0.994 &0.167 &0.667 &0.333\\
CART       &0.169  &0.999 &0.205 &0.584 &0.168 && 0.537 &0.947 &0.154 &0.742 &0.485\\
CTREE      &0.236  &0.998 &0.190 &0.617 &0.234 && 0.523 &0.956 &0.151 &0.739 &0.479\\
Boost.glm  &0.156  &0.926 &0.209 &0.541 &0.082 && 0.457 &0.889 &0.163 &0.673 &0.347\\
N.Bayes    &0.161  &0.998 &0.208 &0.579 &0.159 && 0.195 &0.996 &0.201 &0.596 &0.191\\ \\

& \multicolumn{5}{c}{\scriptsize $\alpha = 0.15$ } & \multicolumn{6}{c}{\scriptsize $\alpha = 0.20$ }\tabularnewline
{\scriptsize Methods} & {\scriptsize Sensibility}& {\scriptsize Specificity}& {\scriptsize Error}& {\scriptsize AUC}& {\scriptsize PSS}& & {\scriptsize Sensibility}& {\scriptsize Specificity}& {\scriptsize Error}& {\scriptsize AUC}& {\scriptsize PSS}\\
\hline
\hline\\
\textbf{PBBC}&0.798 &0.760 &0.230 &0.779 &0.558 && 0.776 &0.789 &0.214 &0.783 &0.565\\
ROSE       &0.847  &0.777 &0.206 &0.812 &0.623 && 0.851  &0.773 &0.208 &0.812 &0.624\\
SMOTE      &0.740  &0.828 &0.194 &0.784 &0.568 && 0.782  &0.806 &0.200 &0.794 &0.588\\
Random.F   &0.779  &0.840 &0.175 &0.809 &0.619 && 0.833  &0.792 &0.198 &0.813 &0.625\\
Boosting   &0.773  &0.871 &0.153 &0.822 &0.644 && 0.838  &0.830 &0.168 &0.834 &0.668\\
CART       &0.495  &0.948 &0.164 &0.722 &0.443 && 0.890  &0.671 &0.275 &0.781 &0.561\\
CTREE      &0.812  &0.815 &0.186 &0.813 &0.627 && 0.835  &0.796 &0.195 &0.815 &0.630\\
Boost.glm  &0.757  &0.762 &0.191 &0.760 &0.519 && 0.858  &0.681 &0.231 &0.770 &0.539\\
N.Bayes    &0.256  &0.991 &0.190 &0.624 &0.247 && 0.293  &0.988 &0.183 &0.640 &0.281\\
\\
\hline
\end{tabular}
\normalsize
\end{center}
\caption{ \footnotesize Sensibility, specificity, error estimation, area under the ROC curve and Pierce score, with different base classification rules, with different proportions ($\alpha$) of the target class, for Adult Data Set. This performances are performed on test sample which distribution is different to distribution of training sample}
\label{table2}
\end{table}

\begin{table}[H]
\begin{center}
\footnotesize
\begin{tabular}{lccccccccccc}
\hline\\
& \multicolumn{5}{c}{\scriptsize $\alpha = 0.03$ } & \multicolumn{6}{c}{\scriptsize $\alpha = 0.07$ }\tabularnewline
{\scriptsize Methods} & {\scriptsize Sensibility}& {\scriptsize Specificity}& {\scriptsize Error}& {\scriptsize AUC}& {\scriptsize PSS}& & {\scriptsize Sensibility}& {\scriptsize Specificity}& {\scriptsize Error}& {\scriptsize AUC}& {\scriptsize PSS}\\
\hline
\hline\\
\textbf{PBBC}&0.999 &0.939 &0.059 &0.970 &0.939 && 0.971 &0.884 &0.110 &0.928 &0.855\\
ROSE      &0.881  &0.930 &0.072 &0.905 &0.810 && 0.979  &0.952 &0.046 &0.965 &0.931\\
SMOTE     &0.798  &0.984 &0.022 &0.891 &0.782 && 0.926  &0.988 &0.016 &0.957 &0.914\\
Boosting  &0.786  &0.960 &0.045 &0.873 &0.746 && 0.973  &0.934 &0.063 &0.954 &0.907\\
Random.F  &0.964  &0.925 &0.074 &0.945 &0.889 && 0.985  &0.943 &0.054 &0.964 &0.928\\
Boost.glm &0.750  &0.969 &0.038 &0.860 &0.719 && 0.994  &0.944 &0.053 &0.969 &0.938\\
CATR      &0.470  &0.984 &0.032 &0.727 &0.454 && 0.847  &0.979 &0.030 &0.913 &0.826\\
CTREE     &0.786  &0.947 &0.058 &0.866 &0.732 && 0.947  &0.950 &0.050 &0.949 &0.897\\
N.Bayes   &0.893  &0.970 &0.032 &0.931 &0.863 && 0.952  &0.970 &0.031 &0.961 &0.921\\
\\

& \multicolumn{5}{c}{\scriptsize $\alpha = 0.15$ } & \multicolumn{6}{c}{\scriptsize $\alpha = 0.30$ }\tabularnewline
{\scriptsize Methods} & {\scriptsize Sensibility}& {\scriptsize Specificity}& {\scriptsize Error}& {\scriptsize AUC}& {\scriptsize PSS}& & {\scriptsize Sensibility}& {\scriptsize Specificity}& {\scriptsize Error}& {\scriptsize AUC}& {\scriptsize PSS}\\
\hline
\hline\\
\textbf{PBBC}&0.981 &0.861 &0.120 &0.921 &0.843 && 0.995 &0.887 &0.081 &0.941 &0.882\\
ROSE      &0.976  &0.958 &0.039 &0.967 &0.935 && 0.989  &0.963 &0.029 &0.976 &0.952\\
SMOTE     &0.957  &0.978 &0.026 &0.967 &0.934 && 0.976  &0.974 &0.026 &0.975 &0.950\\
Boosting  &0.941  &0.961 &0.042 &0.951 &0.901 && 0.972  &0.951 &0.043 &0.962 &0.923\\
Random.F  &0.987  &0.950 &0.044 &0.969 &0.937 && 0.978  &0.967 &0.030 &0.972 &0.945\\
Boost.glm &0.991  &0.932 &0.059 &0.962 &0.924 && 0.962  &0.959 &0.040 &0.961 &0.921\\
CART      &0.905  &0.969 &0.041 &0.937 &0.874 && 0.936  &0.955 &0.051 &0.946 &0.891\\
CTREE     &0.949  &0.925 &0.071 &0.937 &0.874 && 0.976  &0.934 &0.054 &0.955 &0.909\\
N.Bayes   &0.964  &0.971 &0.030 &0.967 &0.935 && 0.988  &0.965 &0.028 &0.976 &0.953\\ \\
\hline
\end{tabular}
\normalsize
\end{center}
\caption{ \footnotesize Sensibility, specificity, error estimation, area under the ROC curve and Pierse score, with different base classification rules, with different proportions ($\alpha$) of the target class, for Breast Cancer Data Set.}
\label{table3}
\end{table}

\begin{table}[H]
\begin{center}
\footnotesize
\begin{tabular}{lccccccccccc}
\hline\\
& \multicolumn{5}{c}{\scriptsize $\alpha = 0.03$ } & \multicolumn{6}{c}{\scriptsize $\alpha = 0.07$ }\tabularnewline
{\scriptsize Methods} & {\scriptsize Sensibility}& {\scriptsize Specificity}& {\scriptsize Error}& {\scriptsize AUC}& {\scriptsize PSS}& & {\scriptsize Sensibility}& {\scriptsize Specificity}& {\scriptsize Error}& {\scriptsize AUC}& {\scriptsize PSS}\\
\hline
\hline\\
\textbf{PBBC}&0.572 &0.854 &0.154 &0.713 &0.426 && 0.550 &0.729 &0.284 &0.639 &0.279\\
ROSE      &0.975  &0.573 &0.415 &0.774 &0.548 && 0.693  &0.806 &0.201 &0.749 &0.499\\
SMOTE     &0.533  &0.971 &0.042 &0.752 &0.504 && 0.620  &0.946 &0.075 &0.783 &0.567\\
Boosting  &0.800  &0.833 &0.168 &0.817 &0.633 && 0.648  &0.773 &0.235 &0.711 &0.421\\
Random.F  &0.544  &0.703 &0.302 &0.623 &0.246 && 0.482  &0.880 &0.146 &0.681 &0.363\\
Boost.glm &0.031  &0.894 &0.132 &0.537 &-0.075 && 0.310 &0.846 &0.189 &0.578  &0.156\\
CART      &0.329  &0.504 &0.501 &0.583 &-0.167 && 0.323 &0.911 &0.128 &0.617  &0.234\\
CTREE     &0.544  &0.861 &0.149 &0.702 &0.405 && 0.501  &0.778 &0.240 &0.639 &0.279\\
N.Bayes   &0.469  &0.854 &0.157 &0.661 &0.323 && 0.658  &0.783 &0.226 &0.720 &0.441\\ \\

& \multicolumn{5}{c}{\scriptsize $\alpha = 0.15$ } & \multicolumn{6}{c}{\scriptsize $\alpha = 0.30$ }\tabularnewline
{\scriptsize Methods} & {\scriptsize Sensibility}& {\scriptsize Specificity}& {\scriptsize Error}& {\scriptsize AUC}& {\scriptsize PSS}& & {\scriptsize Sensibility}& {\scriptsize Specificity}& {\scriptsize Error}& {\scriptsize AUC}& {\scriptsize PSS}\\
\hline
\hline\\
\textbf{PBBC}&0.723 &0.673 &0.320 &0.698 &0.396 && 0.718 &0.733 &0.271 &0.726 &0.451\\
ROSE      &0.695  &0.790 &0.224 &0.742 &0.485 && 0.819  &0.705 &0.263 &0.762 &0.524\\
SMOTE     &0.796  &0.858 &0.151 &0.827 &0.654 && 0.808  &0.855 &0.158 &0.831 &0.663\\
Boosting  &0.760  &0.754 &0.245 &0.757 &0.514 && 0.736  &0.783 &0.230 &0.760 &0.520\\
Random.F  &0.866  &0.708 &0.270 &0.787 &0.573 && 0.802  &0.741 &0.242 &0.772 &0.543\\
Boost.glm &0.509  &0.756 &0.279 &0.633 &0.266 && 0.617  &0.778 &0.268 &0.698 &0.395\\
CART      &0.535  &0.843 &0.202 &0.689 &0.378 && 0.723  &0.716 &0.282 &0.719 &0.439\\
CTREE     &0.693  &0.674 &0.323 &0.684 &0.368 && 0.717  &0.738 &0.268 &0.727 &0.454\\
N.Bayes   &0.784  &0.751 &0.244 &0.768 &0.535 && 0.806  &0.710 &0.263 &0.758 &0.516\\ \\
\hline
\end{tabular}
\normalsize
\end{center}
\caption{ \footnotesize Sensibility, specificity, error estimation, area under the ROC curve and Pierce score, with different base classification rules, with different proportions ($\alpha$) of the target class, for Pima Indians Diabetes Data Set.}
\label{table4}
\end{table}

\section{discussion}
Association rules learning is a well known method in the area of data-mining. It is a research approach for discovering interesting relationships between feature variables in large database. Some algorithms such that linear and logistic regression, k-nearest-neighbour, and Kmeans clusters are "main effect" models and are not able to manage missing values and/or to identify interactions automatically. However the ability to take in account interaction and to manage missing values in building effective predictive models for accurate classification is sometime critical. The main advantages of dealing with association rules learning for classification are : first we  don't need to delete missing values to perform it  and second it can be used to find the best interactions by searching exhaustively all possible combinations of interactions and listing them through  association rules.\\

It appears from the results of Tables \ref{table1}, \ref{table2} that when we are dealing with unbalanced and large dataset the PBBC method is better to use than the Naive Bayes algorithm, the CTREE algorithm, the CART algorithm, Boosting tree algorithm, Boosting generalize linear model algorithm and Random Forest algorithm. The PBBC method produces approximately the same estimation performance than ROSE algorithm and SMOTE algorithm. Moreover when response variable distribution in training sample is different to response variable distribution in test sample the PBBC method is significantly better than SMOTE algorithm when target class occurrence is less than $15\%$.\\
When we are dealing with unbalanced and small dataset the results from Tables \ref{table3} and  \ref{table4} show than the PBBC method produces approximately the same estimations than alternatives methods. 
The main advantage of the PBBC method to others methods such that random forest and boosting methods is that one can present the classifier built by PBBC as an structure tree( see Figure \ref{Figure 1}).\\

In the following table, we sample twenty four classifiers built from unbalanced Adult dataset where the occurrence of the target class is equal to  $\alpha = 0.7\%$. We set the minimum support threshold (Min.sup) from $\{3.5\, 10^{-4}, 4.2 \,10^{-4}, 4.9\, 10^{-4}, 5.6\, 10^{-4}, 6.3 \,10^{-4}, 7.0 \,10^{-4}\}$ and the minimum confidence threshold (Min.conf) from $\{0.02, 0.03, 0.04, 0.05\}$. The process of the PBBC method yields twenty four classifiers  for which the estimations of their performances are presented in the  Table \ref{tab2}. 
 
 
\begin{table}[ht]
\centering
\footnotesize
\begin{tabular}{rrrrrrrr}
  \hline
 & Min.sup & Min.conf & Sensibility & Specificity & Error & AUC & PSS \\ 
  \hline
1 & 3.5 $10^{-4}$ & 0.02 & 0.85 & 0.68 & 0.32 & 0.77 & 0.53 \\
  2 & 3.5 $10^{-4}$ & 0.03 & 0.86 & 0.70 & 0.30 & 0.78 & 0.57 \\ 
  3 & 3.5 $10^{-4}$ & 0.04 & 0.77 & 0.79 & 0.21 & 0.78 & 0.55 \\ 
  4 & 3.5 $10^{-4}$ & 0.05 & 0.72 & 0.84 & 0.16 & 0.78 & 0.56 \\ 
  5 & 4.2 $10^{-4}$ & 0.02 & 0.88 & 0.68 & 0.32 & 0.78 & 0.55 \\ 
  6 & 4.2 $10^{-4}$ & 0.03 & 0.86 & 0.70 & 0.30 & 0.78 & 0.57 \\ 
  7 & 4.2 $10^{-4}$ & 0.04 & 0.77 & 0.79 & 0.21 & 0.78 & 0.55 \\ 
  8 & 4.2 $10^{-4}$ & 0.05 & 0.72 & 0.84 & 0.16 & 0.78 & 0.56 \\ 
  9 & 4.9 $10^{-4}$ & 0.02 & 0.77 & 0.76 & 0.24 & 0.76 & 0.52 \\ 
\rowcolor[gray]{0.7}  10 & 4.9 $10^{-4}$ & 0.03 & 0.85 & 0.74 & 0.26 & 0.80 & 0.59 \\ 
  11 & 4.9 $10^{-4}$ & 0.04 & 0.74 & 0.82 & 0.18 & 0.78 & 0.56 \\ 
  12 & 4.9 $10^{-4}$ & 0.05 & 0.69 & 0.86 & 0.14 & 0.78 & 0.55 \\ 
  13 & 5.6 $10^{-4}$ & 0.02 & 0.77 & 0.73 & 0.27 & 0.75 & 0.49 \\ 
  14 & 5.6 $10^{-4}$ & 0.03 & 0.85 & 0.73 & 0.27 & 0.79 & 0.58 \\ 
  15 & 5.6 $10^{-4}$ & 0.04 & 0.80 & 0.79 & 0.21 & 0.80 & 0.59 \\ 
  16 & 5.6 $10^{-4}$ & 0.05 & 0.62 & 0.88 & 0.12 & 0.75 & 0.49 \\ 
  17 & 6.3 $10^{-4}$ & 0.02 & 0.68 & 0.78 & 0.22 & 0.73 & 0.46 \\ 
  18 & 6.3 $10^{-4}$ & 0.03 & 0.85 & 0.74 & 0.26 & 0.79 & 0.59 \\ 
  19 & 6.3 $10^{-4}$ & 0.04 & 0.79 & 0.79 & 0.21 & 0.79 & 0.58 \\ 
  20 & 6.3 $10^{-4}$ & 0.05 & 0.62 & 0.88 & 0.12 & 0.75 & 0.50 \\ 
  21 & 7.0 $10^{-4}$ & 0.02 & 0.77 & 0.74 & 0.26 & 0.75 & 0.50 \\ 
  22 & 7.0 $10^{-4}$ & 0.03 & 0.84 & 0.74 & 0.26 & 0.79 & 0.58 \\ 
  23 & 7.0 $10^{-4}$ & 0.04 & 0.77 & 0.80 & 0.20 & 0.78 & 0.56 \\ 
  24 & 7.0 $10^{-4}$ & 0.05 & 0.59 & 0.89 & 0.11 & 0.74 & 0.48 \\ 
   \hline
\end{tabular}
\label{tab2}
\caption{Performance estimation of twenty four classifiers using $\alpha = 0.7\%$ as proportion of the rare class both in the training set and in the test dataset.}
\end{table}

\begin{figure}[H] 
\centering
\includegraphics[width=10cm, height=7cm]{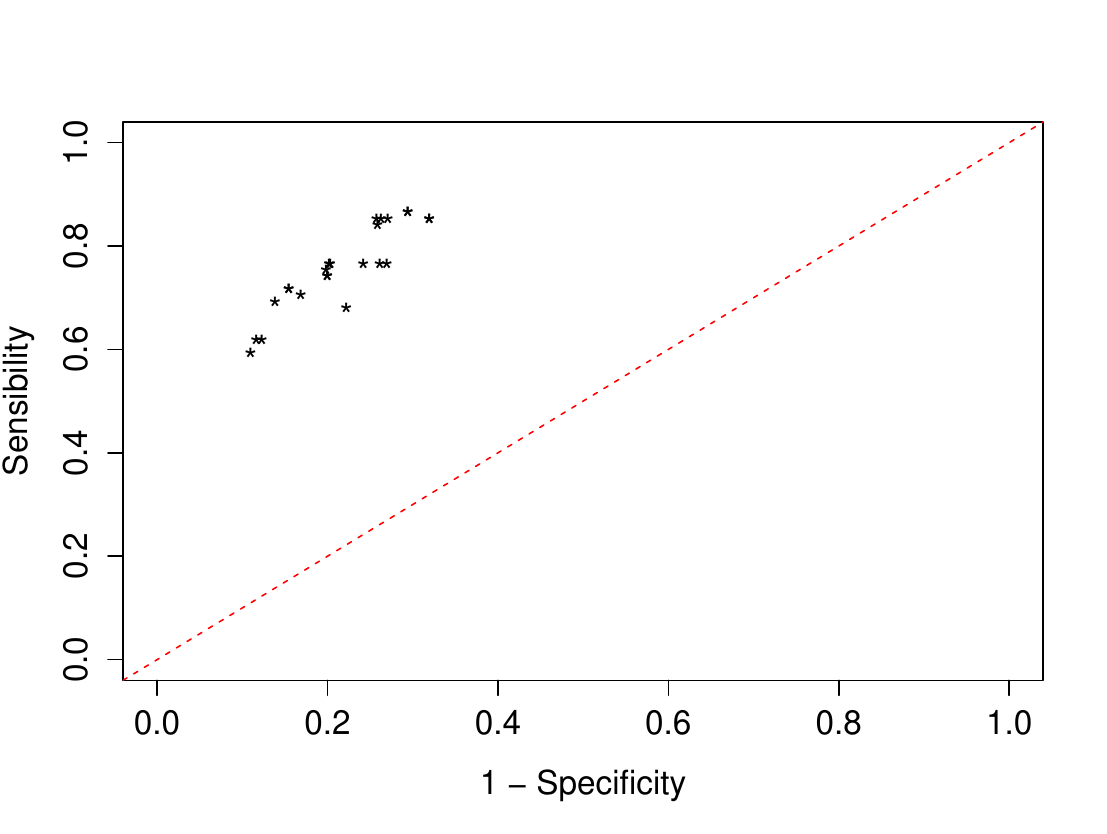} 
\caption{} 
\label{fig1} 
\end{figure} 

The optimal classifier (best sensibility, best specificity and best area under the ROC curve) is given by following learning parameter : Min.sup = $4.9 10^{-4}$ and Min.conf = $0.03$. These above  learning parameters were used to produce an initial set of 271 patterns. The pruning procedure of redundant patterns has allowed to eliminate 121 profiles. The re-evaluation of the performances of the 150 remaining patterns using the validation sample has allowed to identify five patterns whose supports are zero. And the step of selecting relevant patterns has allowed to extract 24 relevant patterns among the 145  non redundant  patterns remaining. The optimal classifier is presented follow as an structure tree (in two parts) that can help to visualise most relevant patterns selected from the sample.
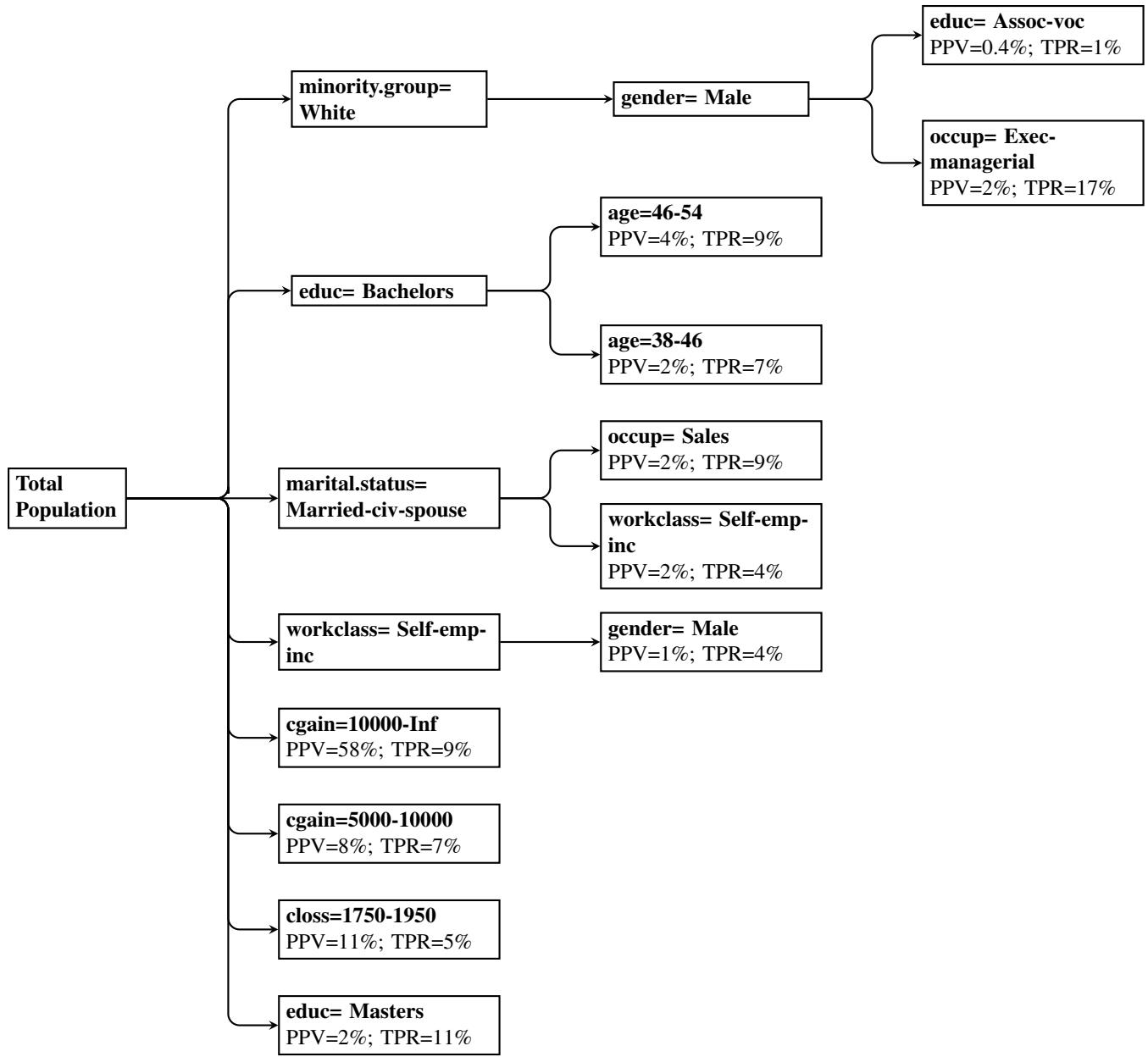
\begin{figure}[H]
\tikzstyle{lien}=[->,>=stealth,rounded corners=5pt,thick]
\tikzset{individu/.style={draw,thick,fill=#1!25},individu/.default={white}}
\begin{center}
\begin{tikzpicture}
\node[individu,draw,text width=2 cm,scale=.8] (I) at (0,3.75) {\textbf{Total\\ Population}};
\node[individu,draw,text width=3.5cm,scale=.8] (A) at (5,10) {\textbf{minority.group= White}};
\node[individu,draw,text width=3.5cm,scale=.8] (A1) at (10,10) {\textbf{gender= Male}};
\node[individu,draw,text width=4cm,scale=.8] (A1_1) at (15,11) {\textbf{educ= Assoc-voc}\\ PPV=0.4\%; TPR=1\%};
\node[individu,draw,text width=4cm,scale=.8] (A1_2) at (15,9) {\textbf{occup= Exec-managerial}\\ PPV=2\%; TPR=17\%};
\node[individu,draw,text width=3.5cm,scale=.8] (B) at (5,7) {\textbf{educ= Bachelors}};
\node[individu,draw,text width=4cm,scale=.8] (B1) at (10,8) {\textbf{age=46-54}\\ PPV=4\%; TPR=9\% };
\node[individu,draw,text width=4cm,scale=.8] (B2) at (10,6) {\textbf{age=38-46}\\ PPV=2\%; TPR=7\%};
\node[individu,draw,text width=4cm,scale=.8] (C) at (5,3.75) {\textbf{marital.status= Married-civ-spouse}};
\node[individu,draw,text width=4cm,scale=.8] (C1) at (10,4.5) {\textbf{occup= Sales}\\ PPV=2\%; TPR=9\%};
\node[individu,draw,text width=4cm,scale=.8] (C2) at (10,3) {\textbf{workclass= Self-emp-inc}\\ PPV=2\%; TPR=4\%};
\node[individu,draw,text width=4cm,scale=.8] (D) at (5,1.5) {\textbf{workclass= Self-emp-inc}};
\node[individu,draw,text width=4cm,scale=.8] (D1) at (10,1.5) {\textbf{gender= Male}\\ PPV=1\%; TPR=4\%};
\node[individu,draw,text width=4cm,scale=.8] (E) at (5,0) {\textbf{cgain=10000-Inf}\\ PPV=58\%; TPR=9\% };
\node[individu,draw,text width=4cm,scale=.8] (F) at (5,-1.5) {\textbf{cgain=5000-10000}\\ PPV=8\%; TPR=7\%};
\node[individu,draw,text width=4cm,scale=.8] (G) at (5,-3) {\textbf{closs=1750-1950} \\ PPV=11\%; TPR=5\% };
\node[individu,draw,text width=4cm,scale=.8] (H) at (5,-4.5) {\textbf{educ= Masters}\\ PPV=2\%; TPR=11\%};
\draw[lien] (I) -| (2.5,4.5)|- (A);
\draw[lien] (A) -- (7.5,10)-- (A1);
\draw[lien] (A1) -| (12.5,10.5)|- (A1_1);
\draw[lien] (A1) -| (12.5,9.5)|- (A1_2);
\draw[lien] (I) -| (2.5,4)|- (B);
\draw[lien] (B) -| (7.5,7.5)|- (B1);
\draw[lien] (B) -| (7.5,6.5)|- (B2);
\draw[lien] (I) -- (2.5,3.75)-- (C);
\draw[lien] (C) -| (7.5,4.12)|- (C1);
\draw[lien] (C) -| (7.5,3.37)|- (C2);
\draw[lien] (I) -| (2.5,2.75)|- (D);
\draw[lien] (D) -- (7.5,1.5)-- (D1);
\draw[lien] (I) -| (2.5,1.85)|- (E);
\draw[lien] (I) -| (2.5,0.75)|- (F);
\draw[lien] (I) -| (2.5,-2.5)|- (G);
\draw[lien] (I) -| (2.5,-3)|- (H);
\end{tikzpicture}
\end{center}
\caption{Presentation of the first part of the structure tree of the optimal classifier among the twenty four classifiers }
\label{Figure 1}
\end{figure}

\begin{figure}[H]
\tikzstyle{lien}=[->,>=stealth,rounded corners=5pt,thick]
\tikzset{individu/.style={draw,thick,fill=#1!25},individu/.default={white}}
\begin{center}
\begin{tikzpicture}
\node[individu,draw,text width=2 cm,scale=.8] (I) at (0,3.5) {\textbf{Total\\ Population}};
\node[individu,draw,text width=3.5cm,scale=.8] (A) at (5,10) {\textbf{hourpw = 50-65}};
\node[individu,draw,text width=3.5cm,scale=.8] (A1) at (10,14.5) {\textbf{gender= Male}};
\node[individu,draw,text width=4cm,scale=.8] (A1_1) at (15,16) {\textbf{age=38-46}\\ PPV=4\%; TPR=11\%};
\node[individu,draw,text width=4cm,scale=.8] (A1_2) at (15,14.5) {\textbf{educ= Bachelors}\\ PPV=5\%; TPR=11\%};
\node[individu,draw,text width=4cm,scale=.8] (A1_3) at (15,12.75) {\textbf{occup= Exec-managerial}\\ PPV=2\%; TPR=5\%};
\node[individu,draw,text width=3.5cm,scale=.8] (A2) at (10,11) {\textbf{minority.group= White}};
\node[individu,draw,text width=4cm,scale=.8] (A2_1) at (15,11) {\textbf{age=38-46}\\ PPV=3\%; TPR=10\%};
\node[individu,draw,text width=4cm,scale=.8] (A3) at (10,9) {\textbf{occup= Prof-specialty}\\ PPV=3\%; TPR=9\%};
\node[individu,draw,text width=4cm,scale=.8] (A4) at (10,7) {\textbf{workclass= Self-emp-inc}\\ PPV=1\%; TPR=1\%};
\node[individu,draw,text width=3.5cm,scale=.8] (B) at (5,3.5) {\textbf{ workclass= Private}};
\node[individu,draw,text width=4cm,scale=.8] (B1) at (10,5) {\textbf{ marital.status= Married-civ-spouse}};
\node[individu,draw,text width=4cm,scale=.8] (B1_1) at (15,5) {\textbf{ age=46-54}\\ PPV=2\%; TPR=9\%};
\node[individu,draw,text width=4cm,scale=.8] (B2) at (10,3.5) {\textbf{ occup= Exec-managerial}};
\node[individu,draw,text width=4cm,scale=.8] (B2_1) at (15,3.5) {\textbf{ gender= Male}\\ PPV=2\%; TPR=10\%};
\node[individu,draw,text width=3.5cm,scale=.8] (B3) at (10,2) {\textbf{ hourpw=42-50}};
\node[individu,draw,text width=4cm,scale=.8] (B3_1) at (15,2) {\textbf{ age=38-46}\\ PPV=2\%; TPR=4\% };
\node[individu,draw,text width=3.5cm,scale=.8] (C) at (5,-2) {\textbf{relation= Husband}};
\node[individu,draw,text width=4cm,scale=.8] (C1) at (10,.5) {\textbf{minority.group= White}};
\node[individu,draw,text width=4cm,scale=.8] (C1_1) at (15,.5) {\textbf{occup= Sales}\\ PPV=1\%; TPR=6\%  };
\node[individu,draw,text width=4cm,scale=.8] (C2) at (10,-1) {\textbf{occup= Prof-specialty}\\ PPV=4\%; TPR=15\%};
\node[individu,draw,text width=4cm,scale=.8] (C3) at (10,-3) {\textbf{occup= Exec-managerial}\\ PPV=3\%; TPR=16\%};
\node[individu,draw,text width=4cm,scale=.8] (C4) at (10,-5) {\textbf{workclass= Self-emp-inc}\\ PPV=2\%; TPR=4\%};
\draw[lien] (I) -| (2.5,7)|- (A);
\draw[lien] (A) -| (7.5,12.25)|- (A1);
\draw[lien] (A1) -| (12.5,15.25)|- (A1_1);
\draw[lien] (A1) -- (12.5,14.5)-- (A1_2);
\draw[lien] (A1) -| (12.5,13.60)|- (A1_3);
\draw[lien] (A) -| (7.5,10.5)|- (A2);
\draw[lien] (A2) -- (12.5,11)-- (A2_1);
\draw[lien] (A) -| (7.5,9.5)|- (A3);
\draw[lien] (A) -| (7.5,8.5)|- (A4);
\draw[lien] (I) -- (2.5,3.5)-- (B);
\draw[lien] (B) -| (7.5,4.5)|- (B1);
\draw[lien] (B1) -- (12.5,5)-- (B1_1);
\draw[lien] (B) -- (7.5,3.5)-- (B2);
\draw[lien] (B2) -- (12.5,3.5)-- (B2_1);
\draw[lien] (B) -| (7.5,2.75)|- (B3);
\draw[lien] (B3) -- (12.5,2)-- (B3_1);
\draw[lien] (I) -| (2.5,1)|- (C);
\draw[lien] (C) -| (7.5,-1.25)|- (C1);
\draw[lien] (C1) -- (12.5,0.5)-- (C1_1);
\draw[lien] (C) -| (7.5,-1.5)|- (C2);
\draw[lien] (C) -| (7.5,-2.5)|- (C3);
\draw[lien] (C) -| (7.5,-3.5)|- (C4);
\end{tikzpicture}
\end{center}
\caption{Presentation of the second part of the structure tree of the optimal classifier among the twenty four classifiers }
\label{Figure 1}
\end{figure}
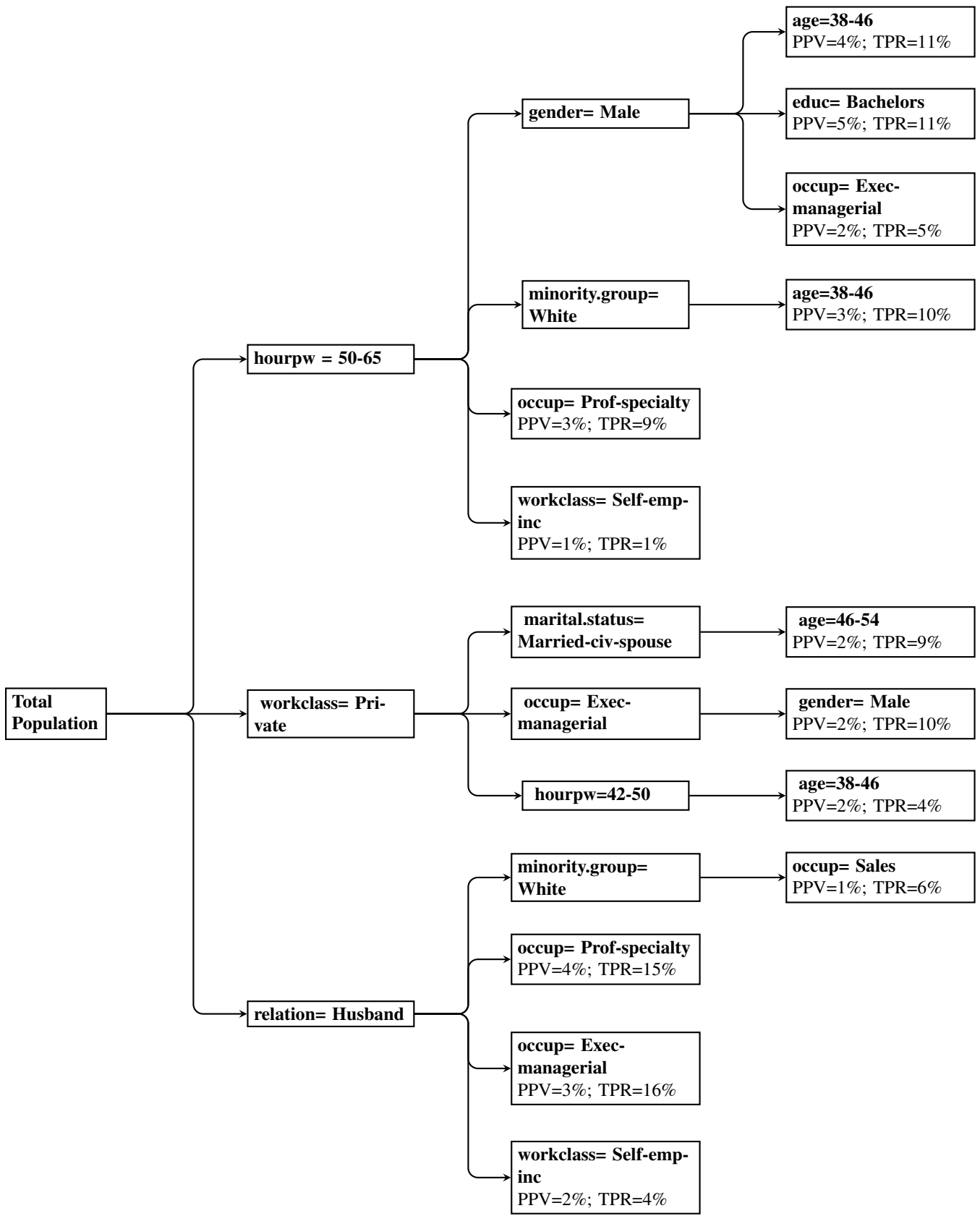

\section{Conclusion}
This paper aims at advocating a methodology to state a binary classification function when dealing with a classification task where the target class is a rare event. Assuming that a large amount of data is available, this goal is achieved by resorting to association rules for exploring the data in order to identify the patterns that are correlated with the target class. Relevant patterns are selected on the basis of their relative risk, their true-positive rates and true-negative rates. The procedure allows to overcome the short-coming of the regression methods which underestimate the conditional probabilities of the occurrence of the target class when the frequency of the instances which belong to this class is very low. Moreover patterns of attributes' interactions which are highly correlated with target class are specified, thus the classification function does not appear like a black-box. Nevertheless one should notice that a stage of data preprocessing is needed before
performing the procedure since it is assumed that the covariates are evaluated on a non-numerical scale. The effectiveness of the proposed method is shown by its application to a real world data related to the study of in-hospital maternal mortality.
\newpage 
\bibliographystyle{acm}
\bibliography{ReferenceArticleFinal}
\end{document}